%% file: www.tex
\documentclass[sigconf]{acmart}

\usepackage{multirow}
\usepackage{pifont}
\usepackage{subcaption}
\newcommand{\cmark}{\checkmark} 
\newcommand{\xmark}{\ding{55}}  

\renewcommand\footnotetextcopyrightpermission[1]{} 
\AtBeginDocument{%
  }

\setcopyright{acmlicensed}
\copyrightyear{2025}
\acmYear{2025}
\acmDOI{XXXXXXX.XXXXXXX}

\usepackage[table]{xcolor}
\definecolor{MyRed}{HTML}{a22f2c}
\definecolor{MyRed2}{HTML}{e22046}
\definecolor{MyYellow}{HTML}{9e7132}
\definecolor{MyGreen}{HTML}{387f61}
\definecolor{MyBlue}{HTML}{2278BF} 
\definecolor{fbApp}{HTML}{ffe4e3}
\definecolor{tabhighlight}{HTML}{e5e5e5}
\newcommand{\rowc}{\rowcolor{fbApp}}

\AtBeginDocument{
  \addtolength{\headheight}{2pt}
  \fancyhead[R]{\textit{Preprint}}
}

\begin{document}



\author{Kuiye Ding}
\email{dingkuiye@ict.ac.cn}
\affiliation{
  \institution{Institute of Computing Technology, Chinese Academy of Sciences}
  \city{Beijing}
  \country{China}
}

\author{Fanda Fan\textsuperscript{\dag}}
\thanks{\textsuperscript{\dag}Corresponding author.}
\email{fanfanda@ict.ac.cn}
\affiliation{
  \institution{Institute of Computing Technology, Chinese Academy of Sciences}
  \city{Beijing}
  \country{China}
}

\author{Zheya Wang}
\email{zheya.wang@durham.ac.uk}
\affiliation{
  \institution{Department of Mathematical Sciences, Durham University}
  \city{Durham}
  \country{UK}
}

\author{Hongxiao Li, Yifan Wang}
\email{lihongxiao19@mails.ucas.ac.cn}
\email{wangyifan2014@ict.ac.cn}
\affiliation{
  \institution{Institute of Computing Technology, Chinese Academy of Sciences}
  \city{Beijing}
  \country{China}
}

\author{Lei Wang, Chunjie Luo}
\email{{wanglei, luochunjie}@ict.ac.cn}
\affiliation{
  \institution{Institute of Computing Technology, Chinese Academy of Sciences}
  \city{Beijing}
  \country{China}
}

\author{Jianfeng Zhan}
\email{zhanjianfeng@ict.ac.cn}
\affiliation{
  \institution{Institute of Computing Technology, Chinese Academy of Sciences}
  \city{Beijing}
  \country{China}
}
\affiliation{
  \institution{University of Chinese Academy of Sciences}
  \city{Beijing}
  \country{China}
}

\renewcommand{\shortauthors}{Kuiye Ding et al.}
\title{KAIROS: Unified Training for Universal Non-Autoregressive \\ Time Series Forecasting}

\begin{abstract}
In the World Wide Web, reliable time series forecasts provide the forward-looking signals that drive resource planning, cache placement, and anomaly response, enabling platforms to operate efficiently as user behavior and content distributions evolve. Compared with other domains, time series forecasting for Web applications requires much faster responsiveness to support real-time decision making. We present \textsc{KAIROS}, a non-autoregressive time series forecasting framework that directly models segment-level multi-peak distributions. Unlike autoregressive approaches, \textsc{KAIROS} avoids error accumulation and achieves just-in-time inference, while improving over existing non-autoregressive models that collapse to over-smoothed predictions. Trained on the large-scale corpus, \textsc{KAIROS} demonstrates strong zero-shot generalization on six widely used benchmarks, delivering forecasting performance comparable to state-of-the-art foundation models with similar scale, at a fraction of their inference cost. Beyond empirical results, \textsc{KAIROS} highlights the importance of non-autoregressive design as a scalable paradigm for foundation models in time series. Code is available at: \url{https://github.com/Day333/Kairos}.


\end{abstract}





\maketitle

\section{Introduction}


The World Wide Web functions as a large-scale, dynamic socio-technical infrastructure in which services must anticipate and respond to shifting content, demand, and user behavior~\cite{huang2025exploiting}. Time series forecasting provides the forward-looking signals required to operate under such non-stationary conditions, leveraging historical telemetry to anticipate future patterns and trends~\cite{10.1145/3485447.3512030, 10.1145/3543507.3583991, 10.1145/3485447.3512037, 10.1145/3589334.3645434,10.1145/3589334.3647982, 10.1145/3589334.3645461, 10.1145/3442381.3450032}. Strong forecasting performance improves user experience and underpins intelligent web services, powering personalized recommendation and multi-peak representation learning~\cite{ning2024debiasing, 10.1145/3543507.3583206}, capacity planning and predictive auto-scaling for microservices~\cite{10.1145/3589334.3645330}, web-scale behavioral and economic modeling~\cite{10.1145/3589334.3645606}, and event-driven financial forecasting~\cite{10.1145/3442381.3450032}. These capabilities position time series forecasting as a foundational component for building adaptive, data-driven platforms across the World Wide Web~\cite{qingqinglong2024unveiling}. Recently, time series foundation models have emerged~\cite{liu2025sundial, shi2024timemoe, goswami2024moment, ansari2024chronos, woo2024moirai, timesfm}, inspired by the pre-training paradigm of large language models, and have demonstrated strong potential in zero-shot forecasting, cross-domain transfer, and long-range dependency modeling.

\begin{figure}[t!]
    \centering
    \includegraphics[width=0.45\textwidth]{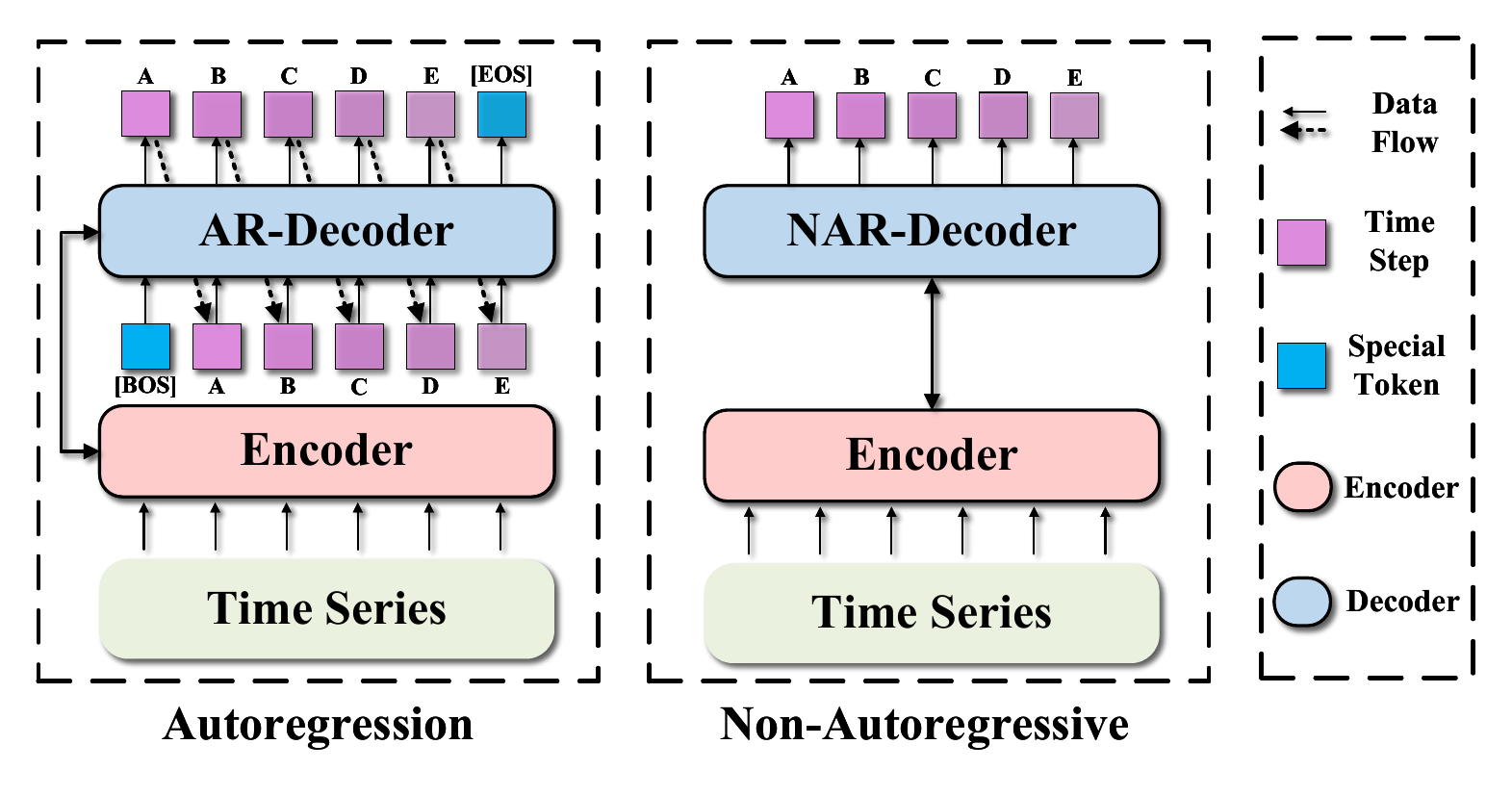}
    \vspace{-10pt}
    \caption{Autoregressive (AR) vs. non-autoregressive (NAR) forecasting. \textbf{Left:} the AR decoder generates each future token/segment sequentially, conditioning on previously produced outputs, which creates strict left-to-right dependencies and inference time that grows with the horizon. \textbf{Right:} the NAR decoder predicts all future segments in parallel from the encoder states in a single pass, removing the sequential dependency and enabling faster decoding.}
    \vspace{-5pt}
    \label{fig:nar} 
\end{figure}

The current landscape of time series modeling exhibits a clear polarization. On one side, direct predictors built upon simple linear layers generate the entire forecast sequence~\cite{Zeng2022AreTE, patchtst, chen2025simpletm} in a single forward pass. This \textit{non-autoregressive (NAR)} strategy~\cite{gui2023nonautoregressive, gu2018nonautoregressive} ensures extremely high inference efficiency, but its constrained representational power makes it inadequate for capturing the complex temporal dynamics of real-world data. On the other side, in pursuit of higher predictive performance, emerging \textit{time series foundation models (TSFMs)} have largely reverted to the \textit{autoregressive (AR)} generative paradigm~\cite{liu2025sundial, shi2024timemoe,ansari2024chronos}. While AR models demonstrate strong modeling power, they also re-expose three intrinsic weaknesses of this paradigm that have long been discussed in sequence generation: \textbf{\textit{(i) slow inference speed}}, as inference time grows linearly with sequence length~\cite{lin-etal-2021-limitations}, which becomes a critical bottleneck in long-horizon scenarios requiring rapid responses; \textbf{\textit{(ii) exposure bias}}, since models rely on teacher forcing with ground-truth labels during training but must depend on their own potentially erroneous predictions at inference, leading to a mismatch that harms generalization~\cite{ranzato2016sequenceleveltrainingrecurrent, gu2018nonautoregressive}; and \textbf{\textit{(iii) error accumulation}}~\cite{ranzato2016sequenceleveltrainingrecurrent}, as small mistakes made early in the inference process can be amplified across subsequent steps, causing predictions to drift away from the ground truth sequence. We illustrate the key difference between AR and NAR forecasting in Figure~\ref{fig:nar}, where AR decoders generate outputs sequentially while NAR decoders predict all future segments in parallel.

\begin{figure}[t!]
    \centering
    \includegraphics[width=0.45\textwidth]{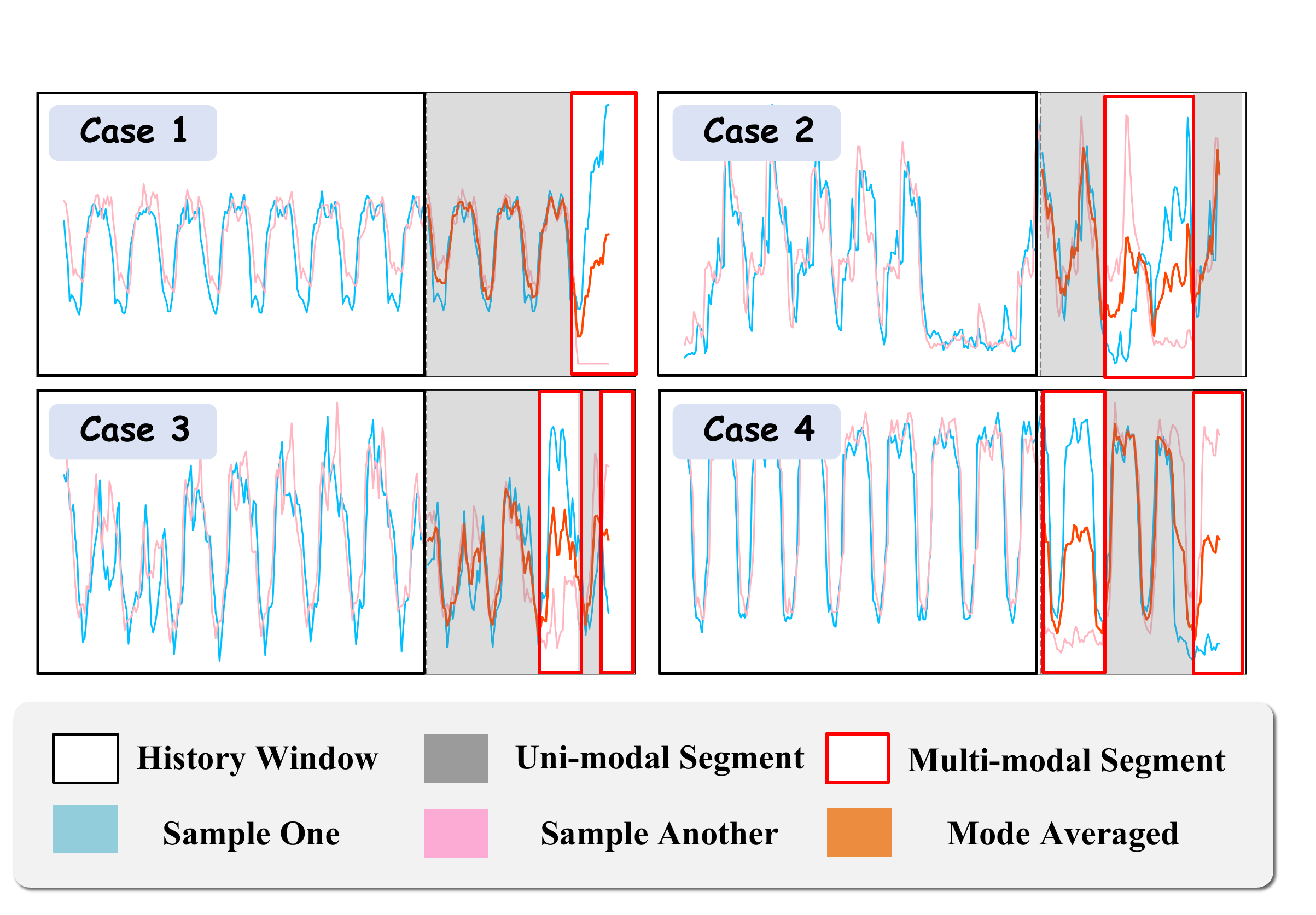}
    \vspace{-5pt}
    \caption{Illustration of four representative cases where the \textit{history windows} exhibit high similarity, yet the corresponding \textit{prediction horizons} differ across segments. Within the forecast, some regions (grey) are relatively uni-modal with consistent sequence, while others (red) are multi-peak with large divergence among plausible futures. In the multi-peak segments, models trained with point-estimation losses tend to produce mode-averaged predictions (orange), leading to over-smoothing or partial mode collapse. Importantly, this effect does not occur uniformly across the horizon but varies segment by segment, motivating our design of segment-wise forecasting to explicitly capture such local variability. These four cases are all from the ECL dataset~\cite{haoyietal-informer-2021}.}
    \vspace{-5pt}
    \label{fig:idea} 
\end{figure}

These shortcomings motivate us to revisit the NAR paradigm, which has been widely regarded as a promising path toward models that combine efficient inference with accurate forecasting, and has already achieved notable success in applications such as recommender systems~\cite{narrec} and machine translation~\cite{gu2018nonautoregressive,gui2023nonautoregressive}. However, applying the NAR paradigm to time series forecasting requires addressing its most fundamental challenge, namely the \textit{Multi-peak Distribution} of target data~\cite{gu2018nonautoregressive}. In time series, this phenomenon is often described as \textit{Temporal Distribution Shift}~\cite{qiu2025duet, brockwell2009time, 10.1145/3459637.3482315}, where similar historical inputs may lead to multiple, equally plausible futures. Unlike a global drift, such variability typically exhibits a \textit{local} nature: certain forecast segments diverge significantly while others remain relatively close, as illustrated in Fig.~\ref{fig:idea}. We therefore characterize this as a segment-level multi-peak distribution problem, and argue that the term multi-peak distribution is more appropriate. This segment-level diversity provides the key motivation for adopting \textit{segment-wise forecasting} in our NAR design. When trained with point-estimation objectives such as MSE, conventional NAR models often suffer from \textit{mode collapse}~\cite{liu2025sundial}, producing averaged outcomes across possible futures and leading to forecasts that are excessively \textit{over-smooth}~\cite{liu2025sundial} and of limited practical value.

To address the above challenges, we propose \textbf{KAIROS}, a high-performance non-autoregressive framework specifically designed to resolve the issue of multi-peak forecasting. Unlike conventional single-model approaches, KAIROS is a complete solution driven by three synergistic mechanisms: \textbf{\textit{(i) Scenario-Aware Generative Experts (SAGE)}}: for future segments with multi-peak distributions, a single predictor tends to collapse different plausible outcomes into an averaged forecast, leading to mode collapse and over-smoothing. SAGE equips each segment with a mixture-of-experts (MoE) prediction head, where different experts specialize in distinct plausible \textit{scenarios}, and a dynamic gating network routes each prediction to the most relevant expert combination. This design explicitly alleviates the multi-peak problem while maintaining parallel generation. \textbf{\textit{(ii) Learnable Exogenous Vectors}}: the multi-peak nature of time series often arises from hidden external factors (e.g., environmental or contextual shifts) that are not directly observable in the input. KAIROS introduces a novel mechanism that combines statistical summaries of the input with a set of learnable \textit{exogenous noise} vectors, effectively approximating these latent external variables to provide unique conditional information for each segment, thereby mitigating the multi-peak challenge. \textbf{\textit{(iii) Segment Causal Residual Noise (SCRN)}}: to mitigate discontinuities across independently generated segments, KAIROS introduces lightweight learnable noise embeddings in a causal residual form. SCRN refines each segment’s prediction by injecting structured noise derived from its immediate predecessor, ensuring that every segment can only condition on past segments while remaining blind to the future. In this way, SCRN inherits the benefits of autoregressive dependency modeling with far lower computational cost, enabling forecasts that evolve smoothly and consistently across segments without reintroducing the inefficiencies of AR models.

Our contributions lie in four aspects:  
\begin{itemize}
\item We propose KAIROS, a non-autoregressive framework for time series forecasting that explicitly addresses the multi-peak challenge through a mixture-of-experts design.
\item We design learnable exogenous vectors to capture latent external factors and SCRN to causally link successive segments, improving temporal coherence without compromising NAR efficiency.
\item Experimentally, KAIROS achieves state-of-the-art  zero-shot performance on multiple forecasting benchmarks with substantially faster inference than autoregressive baselines, and shows clear advantages on long-horizon tasks.
\end{itemize}

\begin{table*}[t!]
\centering
\caption{Key modeling differences between KAIROS and representative TSFMs.}
\resizebox{\textwidth}{!}{
\begin{tabular}{lcccccccccc}
\toprule
\multirow{2}{*}{\centering Method} 
& Kairos & Sundial & Time-MoE & Timer-XL
& Moirai & Moment & LLMTIME
& Chronos& Lag-Llama & TimesFM \\
 & (Ours) & (2025) & (2024) & (2024) & (2024) & (2024) & (2024) & (2024) & (2023) & (2023) \\
\midrule
Non-autoregressive Decoding      & \cmark & \xmark & \xmark & \xmark & \cmark & \cmark & \xmark & \xmark & \xmark & \xmark \\
Adaptive Multi-granularity Patch & \cmark & \xmark & \xmark & \xmark & \xmark & \xmark & \xmark & \xmark & \xmark & \xmark \\
Modeling Multi-peak Distribution& \cmark & \cmark & \xmark & \xmark & \xmark & \xmark & \xmark & \xmark & \xmark & \xmark \\
\bottomrule
\end{tabular}}
\end{table*}

\section{Related Work}
\subsection{Time Series Foundation Models}
Time series foundation models (TSFMs) have achieved remarkable progress in recent years. Most existing TSFMs adopt an AR paradigm~\cite{liu2025sundial, rasul2023lagllama, shi2024timemoe,liu2025timerxl,ansari2024chronos, gruver2023llmtime, timesfm}, which provides strong modeling capacity and has led to significant performance gains. However, AR models also inherit the drawbacks of sequential decoding, including slow inference and exposure bias. A few NAR variants have been explored~\cite{woo2024moirai,goswami2024moment}, but these are primarily encoder-only architectures without a dedicated NAR decoder, and thus cannot fully exploit the advantages of parallel generation. In contrast, our model is designed with an explicit NAR decoding mechanism to address these limitations. Beyond modeling paradigms, current state-of-the-art TSFMs typically rely on either point-wise encoding~\cite{shi2024timemoe, ansari2024chronos,gruver2023llmtime, rasul2023lagllama} or fixed-size patch encoding~\cite{liu2025sundial, liu2025timerxl, woo2024moirai, goswami2024moment, timesfm}. Point-wise representations often lead to prohibitive parameter counts, while fixed patches struggle to capture sufficient temporal variation. To address these limitations, we adopt the adaptive granularity patch  proposed in~\cite{ding2025timemosaictemporalheterogeneityguided}, which enables variable-granularity representations better suited for capturing temporal dynamics in TSFMs.

Time-MoE is a decoder-only \emph{autoregressive} time-series model that scales computation with sparse mixture-of-experts inside the backbone and generates forecasts step by step with point-wise tokenization~\cite{shi2024timemoe}. 
\emph{KAIROS} is \emph{non-autoregressive}: it predicts future segments in parallel. 
Its MoE is placed in scenario-aware segment heads (SAGE) to separate alternative futures rather than to scale an AR backbone; it introduces learnable exogenous vectors to capture latent external drivers; and it employs Segment Causal Residual Noise (SCRN) to refine each segment using only preceding segments, improving temporal coherence without reverting to AR decoding. In short, Time-MoE targets scalable AR pretraining, whereas KAIROS targets segment-level multi-peak and cross-segment linkage within a parallel decoder.

Other related efforts that combine large language models (LLM) with time series forecasting have also shown promising results~\cite{ding2025dualsg,timellm,liu2024autotimes,timecma,zhou2023onefitsall,calf}, but their formulations differ substantially from ours. These approaches typically treat time series as natural language tokens and rely on LLM, whereas our work focuses on designing a dedicated non-autoregressive forecasting framework. Such approaches are therefore not considered TSFMs in this paper, and we do not provide a detailed discussion.

\subsection{Non-Autoregressive Models}
AR models remain the important choice for time series forecasting~\cite{lin2023segrnn, khaldi2023best, stankeviciute2021conformal, alharthi2024xlstmtimelongtermtime}, and recent TSFMs largely adopt the same formulation~\cite{shi2024timemoe, liu2024timer, ansari2024chronos}. While effective, AR decoding introduces exposure bias and stepwise error accumulation~\cite{ranzato2016sequenceleveltrainingrecurrent}, and its latency scales with the forecast horizon, treating short and long horizons uniformly despite different uncertainty and complexity.

NAR generation removes stepwise dependencies and enables parallel prediction. In neural machine translation, NAR substantially reduces latency but faces multimodality and alignment (fertility) issues, latent-variable formulations, knowledge distillation, and iterative refinement~\cite{gu2018nonautoregressive}. Similar ideas have been explored beyond text, e.g., recommendation and reranking~\cite{narrec}, where NAR achieves competitive accuracy with significant efficiency gains~\cite{gui2023nonautoregressive}. Most task-specific full-shot time series forecasting models~\cite{qiu2025duet, hu2025timefilter, stitsyuk2025xpatch, patchtst, Zeng2022AreTE, wang2024timemixer++, chen2024pathformer, qiu2024tfb} adopt a non-autoregressive paradigm, typically relying on a single linear projection, whereas the non-autoregressive formulation of time series foundation models (TSFMs) remains largely underexplored. In time series, however, NAR remains underexplored: existing attempts are predominantly encoder-only or imputation-oriented and lack a dedicated generative NAR decoder, leaving segment-level multimodality and cross-segment dependencies insufficiently addressed. Our work situates a NAR decoder within a forecasting framework that explicitly targets these two gaps: handling multimodal futures at the segment level and preserving temporal coherence without reverting to AR decoding.

\section{Problem Definition}
Time series forecasting is the task of predicting future observations given a history window of past values. For a sequence $\mathbf{x}_{1:T}$, the model takes the most recent $L$ points $\mathbf{x}_{T-L+1:T}$ as context and outputs predictions for the next $H$ steps $\mathbf{x}_{T+1:T+H}$. In the foundation model setting, a TSFM with parameters $\theta$ is pretrained on large-scale corpora to learn a general conditional mapping 
$p_\theta(\mathbf{x}_{T+1:T+H}\mid \mathbf{x}_{T-L+1:T})$. Once trained, the same model can be directly applied in a zero-shot manner on unseen datasets or tasks, i.e., generating forecasts $\hat{\mathbf{x}}_{T+1:T+H}=f_\theta(\mathbf{x}_{T-L+1:T})$ without task-specific finetuning. Objective: learn a universal forecasting model that minimizes the expected loss between predictions and true futures across diverse domains and horizons.

\begin{figure}[!]
\centering
\includegraphics[width=0.9\linewidth]{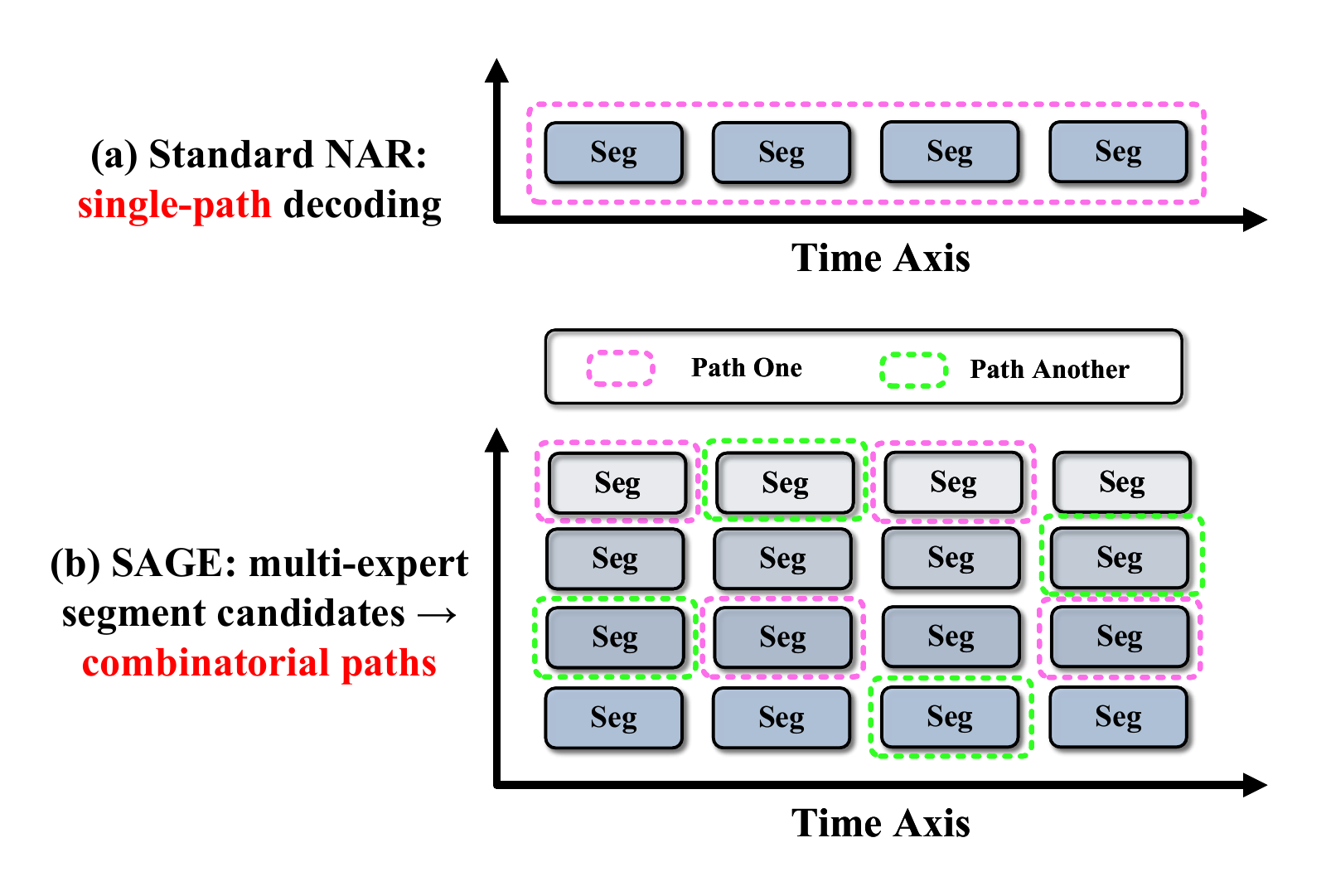}
\vspace{-10pt}
\caption{Illustration of the prediction space with and without Scenario-Aware Generative Experts (SAGE). (i) Standard non-autoregressive decoding outputs one deterministic segment per step, 
resulting in a single prediction path that often collapses to the mean and produces over-smoothed forecasts. (ii) SAGE trains multiple experts to generate diverse candidate segments for each future step. By composing alternative segments along the time axis, the model can explore multiple plausible prediction paths, mitigating mode collapse and improving the fidelity of segment-level forecasts.}
\vspace{-5pt}
\label{fig:sage_paths}
\end{figure}

\begin{figure*}[t!]
    \centering
    \includegraphics[width=0.90\textwidth]{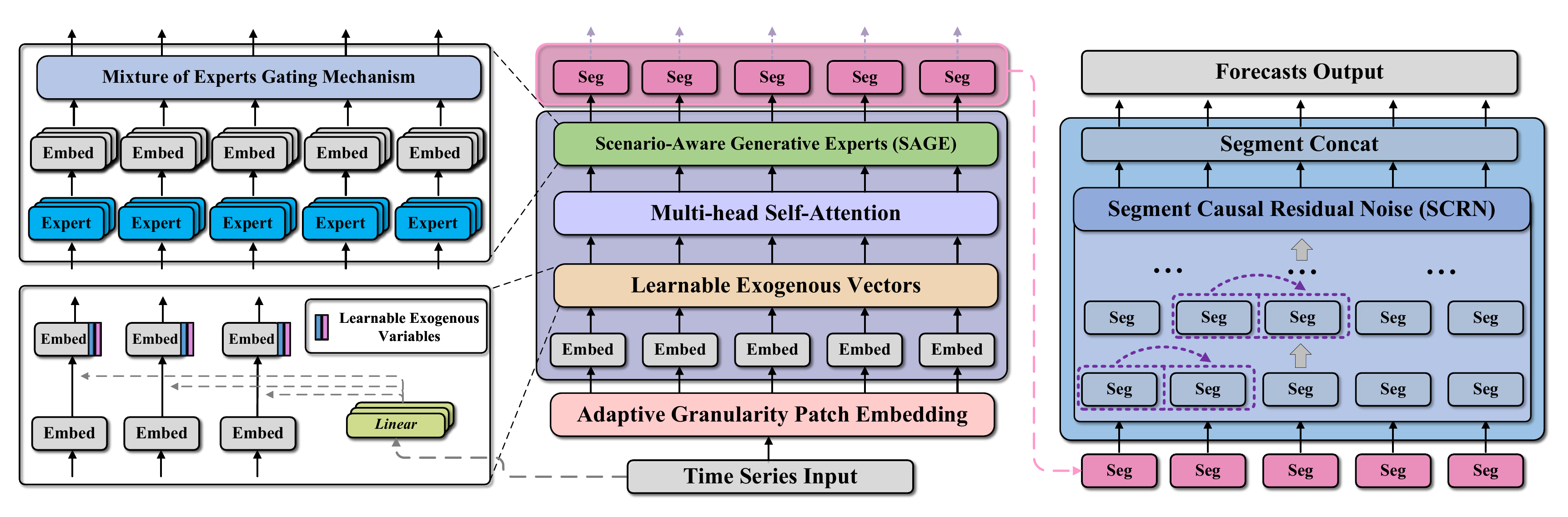}
    \vspace{-10pt}
    \caption{Overview of the proposed \textbf{KAIROS} framework. The model takes time series input and encodes it with adaptive granularity patch embeddings, augmented by learnable exogenous vectors. Scenario-Aware Generative Experts with a mixture-of-experts gating mechanism generate each future segment in parallel. Finally, the Segment Causal Residual FiLM refines segment outputs in a causal manner, linking past and future segments while preserving the efficiency of non-autoregressive decoding.}
    \vspace{-3pt}
    \label{fig:overview}
\end{figure*}

\section{Methodology}

We propose KAIROS, a non-autoregressive framework designed to address segment-level multi-peak in time series forecasting. As shown in Figure~\ref{fig:overview}, KAIROS follows an encoder–decoder architecture with adaptive patch embeddings and learnable tokens to represent historical contexts, while segment-wise mixture-of-experts heads generate multiple future segments in parallel. A causal residual FiLM module then refines predictions across segments, combining the efficiency of non-autoregressive decoding with the temporal awareness of autoregressive models. In the following subsections, we detail each component of the framework, including the encoder design, the scenario-aware generative experts, the learnable exogenous vectors, and the causal refinement module.

\subsection{Scenario-Aware Generative Experts}

A central challenge of non-autoregressive forecasting is to handle multi-peak: for a given historical context, multiple plausible futures may exist. Direct training with pointwise losses often leads to mode averaging and over-smoothed predictions. To mitigate this issue, we introduce Scenario-Aware Generative Experts (SAGE), a segment-wise mixture-of-experts head designed to disentangle alternative futures and generate diverse segment predictions in parallel. As shown in Figure~\ref{fig:sage_paths}, standard non-autoregressive decoding produces a single path that often averages plausible futures, whereas SAGE supplies diverse segment candidates at each step, enabling combinatorial path composition and reducing over-smoothing.

\paragraph{Input representation.}
Let $\mathbf{H}\in\mathbb{R}^{B\times C\times D\times P}$ denote the encoder output for a batch of size $B$, with $C$ channels, hidden dimension $D$, and $P$ patch tokens. For the $k$-th forecast segment of length $S$, we flatten each $(D,P)$ block into $\mathbf{z}_n\in\mathbb{R}^{d}$ where $d=DP$, resulting in $N=BC$ instances.

\paragraph{Routing mechanism.}
Each instance is routed to experts through
\begin{equation}
\mathbf{s}_n=\mathrm{softmax}(\mathbf{W}_g \mathbf{z}_n)\in\mathbb{R}^{E},
\end{equation}
where $\mathbf{W}_g\in\mathbb{R}^{E\times d}$ and $E$ is the number of experts. A sparse gate $\mathbf{g}_n$ is formed by retaining the top-$K$ entries of $\mathbf{s}_n$, which reduces computation and encourages specialization.

\paragraph{Expert predictions.}
Expert $e$ produces a segment prediction by
\begin{equation}
\hat{\mathbf{y}}_{n,e}=\mathbf{W}^{(e)}\mathbf{z}_n+\mathbf{b}^{(e)}\in\mathbb{R}^{S},
\end{equation}
where $\mathbf{W}^{(e)}\in\mathbb{R}^{S\times d}$ and $\mathbf{b}^{(e)}\in\mathbb{R}^{S}$. The final output combines gated experts with a shared lightweight path:
\begin{equation}
\hat{\mathbf{y}}_{n}=\sum_{e=1}^{E} g_{n,e}\,\hat{\mathbf{y}}_{n,e} + \sigma(\mathbf{w}_s^{\top}\mathbf{z}_n)\,\mathbf{W}_s\mathbf{z}_n,
\end{equation}
where $\mathbf{W}_s\in\mathbb{R}^{S\times d}$ and $\mathbf{w}_s\in\mathbb{R}^{d}$ parameterize the shared predictor, and $\sigma$ is the sigmoid activation.

\paragraph{Load balancing.}
To prevent expert collapse, we introduce an auxiliary loss
\begin{equation}
\mathcal{L}_{\mathrm{aux}}=\lambda E \sum_{e=1}^{E} r_e f_e,
\end{equation}
where $r_e=\frac{1}{N}\sum_{n=1}^{N}s_{n,e}$ is the average routing probability and $f_e=\frac{1}{N}\sum_{n=1}^{N}\mathbb{1}\{e\in\mathrm{Top}\text{-}K(\mathbf{s}_n)\}$ is the selection frequency. The coefficient $\lambda$ controls the strength of balancing.

\paragraph{Segment loss.}
For the $k$-th segment, the training loss is
\begin{equation}
\mathcal{L}^{(k)}=\frac{1}{N}\sum_{n=1}^{N}\ell(\mathbf{y}^{(k)}_n, \hat{\mathbf{y}}_{n}),
\end{equation}
where $\mathbf{y}^{(k)}_n\in\mathbb{R}^{S}$ is the ground truth segment and $\ell(\cdot)$ is a forecasting loss. The overall objective for SAGE is $\mathcal{L}^{(k)}+\mathcal{L}_{\mathrm{aux}}$.

By assigning alternative futures to different experts and regularizing expert usage, SAGE alleviates mode collapse and enables the model to generate sharper, more diverse segment-level predictions while retaining the parallelism of non-autoregressive decoding.

\subsection{Learnable Exogenous Vectors}

Non-autoregressive forecasting often suffers from multi-peak: similar historical contexts may correspond to different plausible futures. A key reason is the influence of hidden external factors that are not explicitly recorded in the data (e.g., market shocks, policy interventions, or user behavior changes). To capture this variability, we introduce Learnable Exogenous Vectors (LEV) as a parametric mechanism that injects controllable noise into the prediction process.

For each forecast segment $k\in\{1,\ldots,K\}$, we associate a set of $E$ learnable vectors 
\begin{equation}
\mathbf{v}^{(k)}=\{\mathbf{v}^{(k)}_1,\ldots,\mathbf{v}^{(k)}_E\}, \quad \mathbf{v}^{(k)}_e\in\mathbb{R}^{D},
\end{equation}
where $D$ is the hidden dimension. These vectors are initialized randomly and updated during training. They play the role of \emph{latent exogenous variables}, offering additional conditioning signals beyond the encoder representation.

\paragraph{Integration with encoder outputs.}
Given encoder output $\mathbf{h}^{(k)}\in\mathbb{R}^{N\times D}$ for segment $k$ ($N=BC$ instances), the input to the decoder head is augmented as
\begin{equation}
\tilde{\mathbf{h}}^{(k)} = [\,\mathbf{h}^{(k)};\,\mathbf{v}^{(k)}\,],
\end{equation}
where $[\,\cdot\,;\,\cdot\,]$ denotes concatenation along the token dimension. This augmentation provides the mixture-of-experts head with stochastic context that can explain diverse futures.

By learning to align injected vectors with distributional variations in the data, LEV allows the model to represent latent external influences. This reduces the tendency toward mode averaging and equips the forecasting process with a flexible parametric handle for uncertainty, while keeping inference cost negligible.

\subsection{Segment Causal Residual Noise}

To maintain causal dependencies across adjacent segments while preserving the efficiency of non-autoregressive decoding, we propose \emph{Segment Causal Residual Noise} (SCRN). Unlike FiLM-based modulation that explicitly parameterizes $\gamma$ and $\beta$ via convolution, SCRN introduces lightweight learnable embeddings to inject causal residuals in the form of structured noise.

Let $\mathbf{y}^{(s)}\in\mathbb{R}^{B\times C\times L}$ denote the prediction of the $s$-th segment, where $B$ is the batch size, $C$ the channel dimension, and $L$ the segment length. For each segment index $s\in\{1,\ldots,S\}$, we associate a learnable embedding vector $\mathbf{e}^{(s)}\in\mathbb{R}^{L}$ initialized with Kaiming uniform and scaled by a small factor. When $s>1$, the embedding is broadcast to all batches and channels, and combined with the previous segment prediction $\mathbf{y}^{(s-1)}$ to generate residual noise:
\begin{equation}
\mathbf{n}^{(s)} = \mathbf{y}^{(s-1)} \odot \mathbf{e}^{(s)},
\end{equation}
where $\odot$ denotes element-wise multiplication. The current segment is then refined as
\begin{equation}
\tilde{\mathbf{y}}^{(s)} = \mathbf{y}^{(s)} + \alpha \mathbf{n}^{(s)},
\end{equation}
where $\alpha$ is a small learnable scalar controlling the noise strength. For the first segment $s=1$, no adjustment is applied.

To prevent the embeddings from drifting toward large magnitudes, we regularize them with an $\ell_2$ penalty:
\begin{equation}
\mathcal{L}_{\text{reg}} = \frac{1}{|\mathbf{E}|}\sum_{s=1}^{S}\|\mathbf{e}^{(s)}\|_2^2,
\end{equation}
where $\mathbf{E}$ is the set of all embeddings. This encourages embeddings to stay close to zero, thereby functioning as small perturbations rather than dominant signals.

SCRN can be interpreted as a form of \emph{causal residual noise injection}~\cite{gu-tan-2022-non}: each segment prediction is gently perturbed using information from its immediate predecessor, which stabilizes long-horizon forecasts and reduces abrupt inconsistencies across segments while incurring negligible computational cost.

\subsection{Model Implementation}

KAIROS integrates three major modules: (i) Scenario-Aware Generative Experts for segment-wise parallel generation, (ii) Learnable Exogenous Vectors for capturing latent external variability, and (iii) Segment Causal Residual Noise (SCRN) for causal refinement across segments. In addition, the encoder employs \emph{adaptive patch embedding} to represent histories at variable granularities. All components are trained jointly in an end-to-end fashion.

\paragraph{Forecasting loss.}
Given ground truth segments $\{\mathbf{y}^{(k)}\}_{k=1}^{K}$ and predictions $\{\hat{\mathbf{y}}^{(k)}\}_{k=1}^{K}$ after SCRN refinement, the primary forecasting loss is
\begin{equation}
\mathcal{L}_{\text{pred}} = \frac{1}{K}\sum_{k=1}^{K} \ell\big(\mathbf{y}^{(k)},\,\hat{\mathbf{y}}^{(k)}\big),
\end{equation}
where $\ell(\cdot,\cdot)$ is a pointwise criterion that are mean squared error (MSE) or mean absolute error (MAE) . 

\paragraph{Expert balancing.}
SAGE introduces an auxiliary loss to avoid expert collapse and encourage balanced routing:
\begin{equation}
\mathcal{L}_{\text{aux}} = \lambda_{\text{aux}} E \sum_{e=1}^{E} r_e f_e,
\end{equation}
where $r_e$ and $f_e$ denote the average routing probability and selection frequency of expert $e$, $E$ is the number of experts, and $\lambda_{\text{aux}}$ is a balancing weight.

\paragraph{Patch budget loss.}
To prevent adaptive patching from degenerating into a single granularity, we regularize the frequency of patch selections. Let $p_m$ denote the empirical probability of choosing patch length $m$ across training samples and $u_m=1/M$ the uniform target distribution over $M$ candidate patch sizes. The budget loss is defined as
\begin{equation}
\mathcal{L}_{\text{budget}} = \lambda_{\text{budget}} \sum_{m=1}^{M} \big(p_m - u_m\big)^{2},
\end{equation}
where $\lambda_{\text{budget}}$ controls the strength of regularization. This encourages the model to explore all granularities during training.

\paragraph{objective.}
The total loss is the weighted sum
\begin{equation}
\mathcal{L} = \mathcal{L}_{\text{pred}} + \mathcal{L}_{\text{aux}} + \mathcal{L}_{\text{budget}}.
\end{equation}
LEV and SCRN are trained implicitly through $\mathcal{L}$ without additional terms.

\section{Experiments}
This section investigates the following key research questions through extensive empirical studies:

\begin{itemize}
    \item \textbf{RQ1:} How well does {KAIROS} generalize in zero-shot forecasting scenarios compared with state-of-the-art autoregressive and non-autoregressive baselines?
    \item \textbf{RQ2:} Does the non-autoregressive formulation of {KAIROS} lead to measurable inference speedup while preserving forecasting performance?
    \item \textbf{RQ3:} What is the impact of varying the design parameters of {KAIROS}, such as the number of experts in SAGE, the dimensionality of learnable exogenous vectors, and the capacity of the SCRN module?
\end{itemize}

\subsection{Dataset and Experimental Settings}

\paragraph{Datasets.} Following TimeMoE~\cite{shi2024timemoe}, we evaluate our model on six widely used benchmarks: ETTh1, ETTh2, ETTm1, ETTm2, Weather, and GlobalTemp. For pre-training, we leverage BLAST~\cite{blast}, a large-scale dataset designed for universal time series modeling. BLAST contains 321 billion observations drawn from diverse sources including CMIP6~\cite{CMIP6} (32.5\%), ERA5~\cite{ERA5} (30.0\%), WeatherBench~\cite{weatherbenchmark} (25.7\%), Buildings\_900K~\cite{emami2023buildingsbench} (6.9\%), and over 300 additional public datasets~\cite{blast} (4.9\%), covering domains such as climate, meteorology, and energy. To avoid data leakage, benchmark datasets used for downstream evaluation (e.g., ETTh/ETTm, Weather, GlobalTemp) are explicitly excluded from BLAST during pre-training. Consequently, other datasets such as ECL, Traffic, and PEMS are not included in our evaluation since they appear in the pre-training corpora of several existing TSFMs, making fair comparison difficult. Moreover, although GIFT-Eval~\cite{aksu2025gifteval} provides a comprehensive evaluation protocol, we do not adopt it here because many of its benchmark datasets overlap with our pre-training corpus, which could compromise evaluation independence.

\paragraph{Baseline.} We compare against pre-trained foundation time series models, including TimeMoE~\cite{shi2024timemoe}, MOIRAI~\cite{woo2024moirai}, Chronos~\cite{ansari2024chronos}, TimesFM~\cite{timesfm}, and Moment~\cite{goswami2024moment}, using the parameter settings as specified in their respective publications. For fairness, we omit TimesFM on the Weather dataset that was included in its pretraining corpus. Sundial and Timer-XL did not report benchmark results on the GlobalTemp dataset.

\paragraph{Evaluation Metrics.} Following previous works, we use Mean Squared Error (MSE) and Mean Absolute Error (MAE) metrics to assess the performance.

\paragraph{Implementation Details.} 
For the ablation studies, we set the number of encoder layers to 1, the input length to 512, and the prediction length to 720. The adaptive patch granularity list is fixed as [8, 16, 32, 64]. We train for 1 epoch with 8 attention heads and a batch size of 32 on a single NVIDIA RTX 3090 GPU. For the experiments reported in Table~\ref{tab:main_results}, we set the batch size to 1024. The training is conducted on 8 NVIDIA V100-48G GPUs and requires approximately 24 hours to complete. All other hyperparameters follow the settings described in previous work~\cite{timesnet,liu2023itransformer,ding2025timemosaictemporalheterogeneityguided}. The results of other baselines in Table~\ref{tab:main_results} come from \cite{liu2025sundial, shi2024timemoe}.

\input{Table/main_results}

\begin{figure}[t!]
\centering
\includegraphics[width=0.9\linewidth]{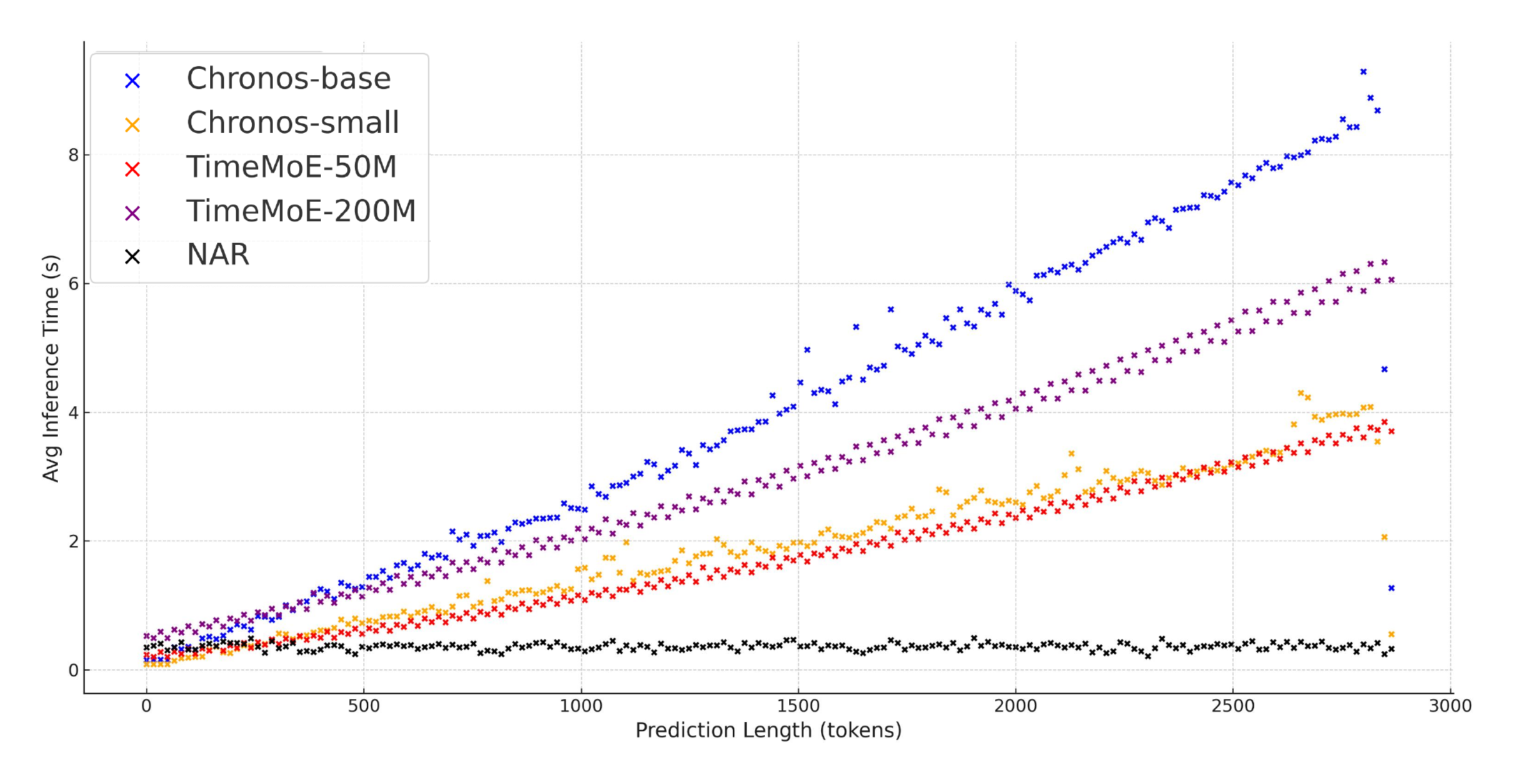}
\vspace{-10pt}
\caption{Comparison of inference times across different time-series foundation models. The x-axis denotes the prediction length (tokens), and the y-axis shows the average inference time (seconds). Results include Chronos-base, Chronos-small, TimeMoE-50M, TimeMoE-200M, and our proposed NAR model.}
\vspace{-5pt}
\label{fig:inference_time}
\end{figure}

\subsection{Zero-shot across Benchmarks (RQ1)}

We evaluate zero-shot generalization on six benchmarks (ETTh1, ETTh2, ETTm1, ETTm2, Weather, GlobalTemp) and four horizons \{96, 192, 336, 720\}, using MSE and MAE without task-specific fine-tuning. The model is pre-trained on BLAST with evaluation datasets excluded to avoid leakage. Table~\ref{tab:main_results} reports comparisons against representative autoregressive and non-autoregressive TSFMs.

Results show that \textsc{KAIROS} achieves performance broadly comparable to strong baselines across datasets and horizons. With 130M parameters, it is on par with \textsc{Sundial}\(_{\text{base}}\) (128M) and \textsc{TimeMoE-50M} (113M), while smaller than \textsc{Chronos}\(_{\text{base}}\) (205M) and \textsc{TimeMoE} (453M). Our Kairos model demonstrates remarkable efficiency: it can be trained with only a \textbf{single NVIDIA RTX 3090 GPU}, or completed within one day on 8 NVIDIA V100-48G GPUs for the full training. In contrast, Sundial requires \textbf{32 A100 GPUs} and TimeMoE requires \textbf{128 A100 GPUs with 10 days} of training. This parity is consistent with our design goal: to demonstrate that a non-autoregressive framework can match the accuracy of established foundation models under strict zero-shot evaluation, while preparing the ground for efficiency gains discussed in RQ2. The competitiveness of \textsc{KAIROS} is therefore both expected and desirable, confirming that explicit modeling of segment-level multi-peak distributions can yield accuracy comparable to larger autoregressive models, without trading off inference efficiency.

\subsection{Inference Efficiency (RQ2)}

A central motivation for adopting the NAR formulation is to remove the strict left-to-right dependency of AR decoding. As shown in Fig.~\ref{fig:inference_time}, the inference time of KAIROS remains nearly constant across different prediction lengths, demonstrating its scalability advantage. In contrast, AR-based TSFMs such as Chronos and TimeMoE exhibit linearly increasing latency as the horizon grows, since each step requires conditioning on previously generated outputs. This stepwise generation becomes a critical bottleneck in long-horizon forecasting, limiting their applicability in real-time scenarios.


When compared with Sundial~\cite{liu2025sundial}, a different behavior is observed. Sundial exhibits staircase-like transitions in inference time, with abrupt jumps at horizons that are multiples of 720 tokens. This arises from its multi-patch prediction design, where each block outputs a fixed length of 720, leading to block-wise parallel prediction for horizons shorter than 720. As a result, its runtime often resembles that of non-autoregressive models. Moreover, Sundial incorporates system-level optimizations such as FlashAttention, KV caching, and shared condition, which further improve efficiency. Consequently, Sundial is substantially faster than conventional AR baselines and can achieve near NAR-like runtime for horizons within 720 tokens. However, due to its reliance on multi-patch generation rather than strict stepwise decoding, Sundial no longer fits the definition of a purely autoregressive model. For this reason, we do not include Sundial in the direct AR-vs-NAR runtime comparison in Figure~\ref{fig:inference_time}.

In summary, the non-autoregressive formulation of KAIROS yields consistent and efficient inference across horizons, while conventional AR TSFMs suffer from increasing latency. Even with advanced optimizations, AR models such as Sundial only partially alleviate this limitation under specific settings, confirming the inherent advantage of NAR decoding for time series foundation models.





\begin{figure}[t]
  \centering
  \begin{subfigure}[t]{0.49\linewidth}
    \centering
    \includegraphics[width=\linewidth]{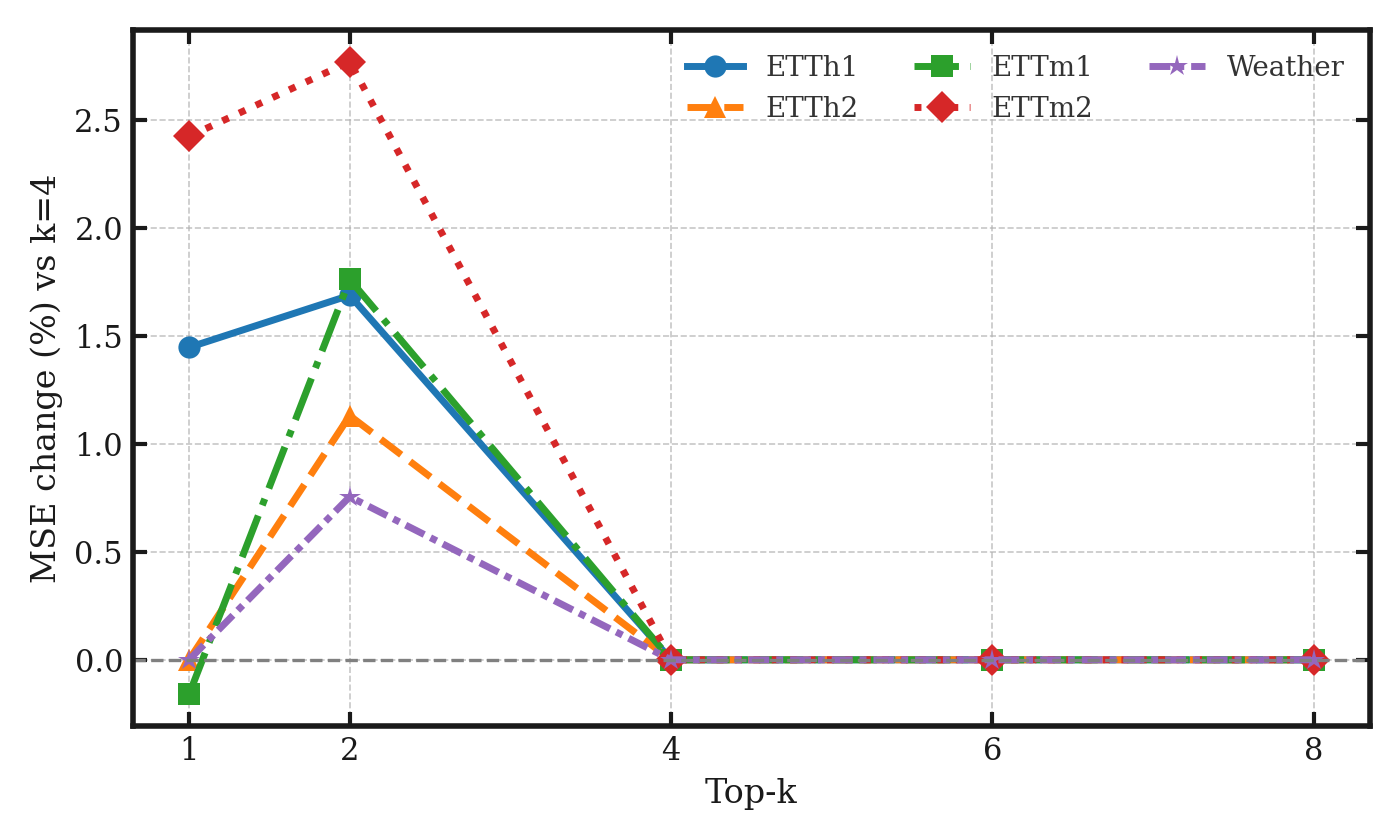}
    \caption{MSE $\Delta$\% (k=4 ref.)}
    \label{fig:rel_mse}
  \end{subfigure}\hfill
  \begin{subfigure}[t]{0.49\linewidth}
    \centering
    \includegraphics[width=\linewidth]{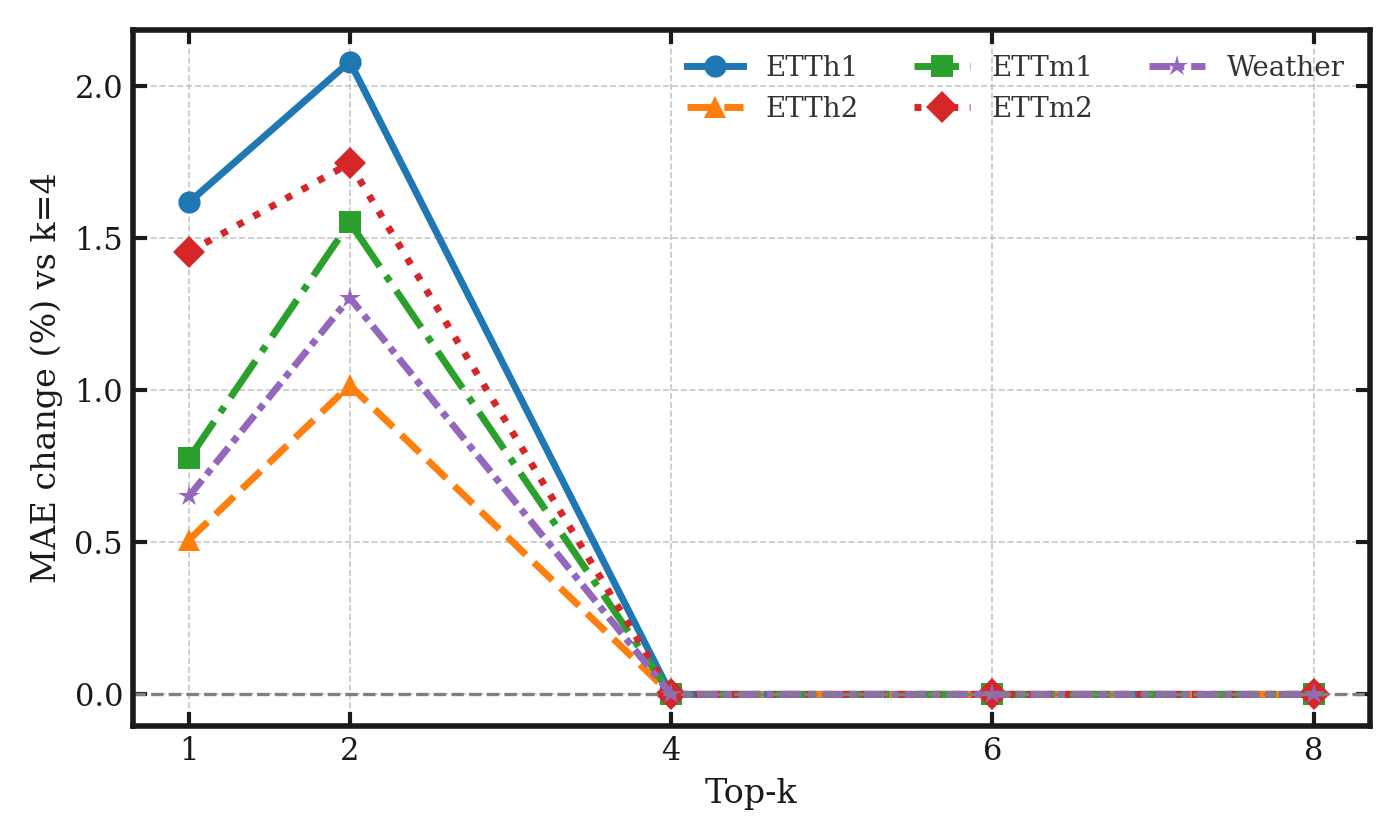}
    \caption{MAE $\Delta$\% (k=4 ref.)}
    \label{fig:rel_mae}
  \end{subfigure}
  \vspace{-3pt}
  \caption{Relative performance change across datasets when varying Top-$k$ (experts $E{=}8$). Values are percentage changes w.r.t.\ $k{=}4$; gray dashed line marks 0\%.}
  \label{fig:rel_change_side_by_side}
  \vspace{-5pt}
\end{figure}

\subsection{Effects of Key Design Parameters (RQ3)}

Due to computational constraints, all ablation experiments in this subsection are conducted on a 1\% subset of the BLAST corpus. This reduced setting still preserves sufficient diversity to examine the effect of different architectural choices, while allowing controlled comparisons of model variants.

\paragraph{Segment Cross-Attention Decoder.} 
To further investigate causal refinement strategies, we implement \emph{Segment Cross-Attention Decoder} (SCAD) as a comparison to SCRN. Instead of injecting lightweight residual noise, SCAD employs a cross-attention mechanism to explicitly condition each forecast segment on its immediate predecessor. Concretely, the first segment is predicted directly from the hidden representation, while subsequent segments are generated by querying their predecessors through a multi-head attention module. Although this design captures richer dependencies across adjacent segments, it introduces additional computational cost, as each new segment requires a full attention pass. We therefore use SCAD primarily as a controlled baseline to assess whether the added complexity of attention-based refinement yields tangible benefits over the lightweight noise-based approach.

\input{Table/ablation_short}

\paragraph{Ablation study of core modules.} We conduct an ablation study to quantify the contribution of each proposed component, as summarized in Table~\ref{tab:ablation_avg}. The \textit{Baseline} corresponds to a simplified encoder–decoder with adaptive patch embeddings but without any of the proposed modules. Adding LEV does not consistently improve performance; in fact, in most cases it slightly degrades accuracy. A possible reason is that the injected exogenous vectors may introduce more stochastic noise than useful conditioning signals, thereby increasing the risk of overfitting rather than capturing genuine latent factors. This suggests that while exogenous variability is an important consideration, learning it implicitly is non-trivial and may require more sophisticated regularization. In contrast, SAGE proves to be the most effective component, yielding clear improvements across datasets (e.g., substantial gains on ETTm2). This validates our claim that segment-wise mixture-of-experts is crucial for handling multi-peak distributions and avoiding mode collapse, making it the main driver of KAIROS’s forecasting performance. Adding SCRN on top of LEV and SAGE provides only marginal benefits and in some cases even offsets the gains. This may be due to the fact that SAGE already provides strong segment-level predictions, and SCRN’s residual correction sometimes introduces additional noise rather than meaningful refinement. Although SCRN is designed to reduce discontinuities, prior work has pointed out that temporal coherence can often emerge implicitly from self-attention representations without explicit local corrections~\cite{wang2025fredf}. In this sense, SCRN plays more of a stabilizing role than a decisive factor for accuracy. Finally, SCAD, the cross-attention alternative, achieves competitive results on Weather but generally underperforms SCRN, reflecting its higher computational cost and tendency to overfit local dependencies. Overall, the ablation confirms that {SAGE is the dominant contributor}, while LEV and SCRN provide auxiliary but less reliable gains. The full KAIROS model (LEV + SAGE + SCRN) is adopted as the default configuration in subsequent experiments, though the analysis highlights the importance of segment-level expert modeling as the key source of improvement.

\paragraph{Hyperparameter Sensitivity.} In Fig.~\ref{fig:rel_change_side_by_side}, we ablate the gating sparsity by varying the Top-$k$ (with $E{=}8$ total experts) on ETTh1, ETTh2, ETTm1, ETTm2 and Weather. Both MSE and MAE curves generally improve from $k{=}1$ to $k{=}4$, and then plateau or slightly degrade for $k{\ge}6$. Averaging over horizons and datasets, Top-$k{=}4$ achieves the lowest error, confirming it as the optimal setting for our MoE configuration, see details in Appendix Table~\ref{tab:moe_topk_avg}.

\paragraph{Limitations and outlook.} The ablation results make clear that the strongest improvements of \textsc{KAIROS} originate from the segment-wise expert design, while the auxiliary modules show mixed effectiveness. The learnable exogenous vectors often degrade performance, which we attribute to the difficulty of distinguishing informative latent factors from stochastic perturbations. In practice, the vectors may introduce additional noise or encourage the model to overfit, especially in zero-shot scenarios where the alignment between pre-training and downstream distributions is fragile. The Segment Causal Residual Noise module also demonstrates limited benefit. Although its purpose is to reduce discontinuities across segments, its corrective signals can sometimes conflict with the already specialized predictions generated by the experts. In those cases, the adjustments may act more like disturbances than refinements. This phenomenon is consistent with prior observations in sequence modeling, where implicit correlations between consecutive predictions already provide a degree of temporal coherence without explicit correction.

We acknowledge that our current design is therefore not a fully effective solution. Nonetheless, these results do not undermine the central message of our work. On the contrary, they emphasize two essential lessons. First, non-autoregressive forecasting is crucial for building foundation models that scale to web-scale time series data while meeting the demands of real-time inference. Second, introducing localized causal refinement remains a promising idea, even though our present implementation yields only modest improvements. This direction is inspired by related progress in machine translation, where similar mechanisms have been shown to mitigate error accumulation. Our findings highlight that the challenge is not in the validity of the idea but in its execution: future work must focus on designing refinement strategies that reinforce rather than conflict with expert predictions. By recognizing these limitations explicitly, we aim to make clear where the most pressing challenges lie and to underline the importance of pursuing efficient non-autoregressive approaches as the foundation for the next generation of time series forecasting models.

\section{Conclusion}

In this work, We introduce \textsc{KAIROS}, a non-autoregressive time series foundation model that explicitly tackles the segment-level multi-peak nature of future distributions. By combining adaptive patch embeddings with segment-wise generative experts, \textsc{KAIROS} achieves diverse predictions and efficient parallel decoding. Pre-trained on BLAST and evaluated in zero-shot settings, \textsc{KAIROS} delivers performance comparable to leading autoregressive and non-autoregressive models, while offering significant inference efficiency gains. Although some auxiliary modules show mixed effectiveness, our study underscores the necessity of non-autoregressive design for scalable time series forecasting and highlights promising directions for improving causal refinement and exogenous variability modeling.



\begin{acks}
This work was supported by the Innovation Funding of Institute of Computing Technology, Chinese Academy of Sciences under Grant No. E461070.
\end{acks}

\bibliographystyle{ACM-Reference-Format}
\bibliography{www}

\clearpage

\appendix

\section{Dataset}

\input{Appendix/dataset}

\section{Sensitivity Analysis }

\paragraph{Exogenous vector size.}
Table~\ref{tab:exo_noise} reports the effect of varying the dimensionality $d_{\text{exo}}$ of the learnable exogenous vectors. Across all datasets, smaller dimensions ($d_{\text{exo}}{=}1$ or $2$) lead to the most stable performance, while further enlarging the vector size generally results in degraded accuracy. This trend suggests that the injected vectors introduce more noise than informative conditioning when their capacity is too large, thereby weakening generalization. In practice, a compact design is sufficient to capture latent variability, whereas higher-dimensional exogenous vectors risk amplifying stochastic fluctuations and overfitting to training data.

\paragraph{Segment length.}
Table \ref{tab:segment_length} reports the sensitivity of KAIROS to different segment lengths across five benchmark datasets. Overall, the results indicate that performance remains stable when varying the segment size from 8 to 48, with only marginal fluctuations in MSE and MAE. On ETTh1 and ETTh2, shorter segments (8 or 16) achieve slightly better accuracy, suggesting that finer-grained decomposition can better capture local dynamics. In contrast, on high-frequency datasets such as ETTm1 and ETTm2, longer segments exhibit comparable or even improved performance, likely due to their ability to aggregate short-term noise. For the Weather dataset, the choice of segment length shows minimal impact, highlighting the robustness of the proposed framework to this hyperparameter. These findings confirm that KAIROS is not overly sensitive to segment size, maintaining stable forecasting accuracy across diverse temporal resolutions.
\input{Table/seg}
\input{Table/exo_noise}

\section{Complexity Analysis}

Let $K$ be the prediction horizon, $S$ the number of forecast segments, $L$ the segment length, and $d$ the hidden dimension.  

\paragraph{Autoregressive decoding.}  
AR models generate one step at a time, yielding a total complexity of
\begin{equation}
\mathcal{O}_{\text{AR}} = \mathcal{O}(K \cdot d^2),
\end{equation}
which scales linearly with $K$ and limits parallelism.

\paragraph{Non-autoregressive decoding.}  
KAIROS predicts $S{=}K/L$ segments in parallel, each via a mixture-of-experts head:
\begin{equation}
\mathcal{O}_{\text{NAR}} = \mathcal{O}(S \cdot d^2).
\end{equation}
Thus the cost depends on $S$ rather than $K$, and all segments can be generated in a single forward pass.

\paragraph{Parallelism.}  
This design makes inference nearly independent of horizon length, enabling substantial speedup over AR decoding while retaining segment-level modeling flexibility.

\section{Formalization of Multi-Peak Distribution}
\label{app:multimodal}

In forecasting, the conditional distribution of future $\mathbf{Y}$ given history $\mathbf{X}$ is often \emph{multi-peak}:
\begin{equation}
p(\mathbf{Y}\mid\mathbf{X}) = \sum_{m=1}^{M} \pi_m(\mathbf{X}) \, p_m(\mathbf{Y}\mid\mathbf{X}),
\end{equation}
where different modes $p_m$ correspond to distinct but plausible futures.  

Conventional NAR models trained with pointwise losses approximate the expectation
\begin{equation}
\hat{\mathbf{Y}} \approx \mathbb{E}_{p(\mathbf{Y}\mid\mathbf{X})}[\mathbf{Y}],
\end{equation}
leading to over-smoothed predictions. This motivates segment-wise expert modeling to preserve diverse futures.

\begin{table*}[htbp]
  \caption{Detailed dataset descriptions. Dataset sizes are listed as (Train, Validation, Test).}
  \label{tab:dataset}
  \centering
  {\scriptsize
  \begin{tabular}{c|l|c|c|c|c|c|c}
    \toprule
    Tasks & Dataset & Dim & Series Length & Dataset Size & Frequency & Forecastability$\ast$ & Information \\
    \midrule
     & ETTm1 & 7 & \{96, 192, 336, 720\} & (34465, 11521, 11521)  & 15min & 0.46 & Temperature \\
    \cmidrule{2-8}
     & ETTm2 & 7 & \{96, 192, 336, 720\} & (34465, 11521, 11521)  & 15min & 0.55 & Temperature \\
    \cmidrule{2-8}
     Long-term & ETTh1 & 7 & \{96, 192, 336, 720\} & (8545, 2881, 2881) & Hourly & 0.38 & Temperature \\
    \cmidrule{2-8}
     Forecasting & ETTh2 & 7 & \{96, 192, 336, 720\} & (8545, 2881, 2881) & Hourly & 0.45 & Temperature \\
    \cmidrule{2-8}
     & Weather & 21 & \{96, 192, 336, 720\} & (36792, 5271, 10540)  & 10min & 0.75  & Weather \\
     \cmidrule{2-8}
     & Global Temp & 1000 & \{96, 192, 336, 720\} & (12280, 1755, 3509)  & Hourly & 0.78  & Temperature \\
    \bottomrule
  \end{tabular}
  }
\end{table*}

\input{Table/ablation}

\input{Table/topk}

\end{document}

%% file: Table/main_results.tex
\begin{table*}[t!]
\setlength{\abovecaptionskip}{.2 cm}
\caption{
Zero-shot Results. Lower MAE and MSE values indicate superior performance. The symbols $s$, $b$, and $l$ represent the small, base, and large versions, respectively. Models with top-3 performance are highlighted in {\textbf{bold}}.
}
\setlength\tabcolsep{1.2pt}
\resizebox{1\linewidth}{!}{
\begin{tabular}{rr|cc|cc|cc|cc|cc|cc|cc|cc|cc|cc|cc|cc}
\toprule[1.5pt]
\multicolumn{2}{l}{\multirow{2}{*}{Method}}  & \multicolumn{2}{|c|}{\textbf{KAIROS}}  & \multicolumn{2}{l|}{\textbf{Sundial}$_{b}$} & \multicolumn{2}{l|}{\textbf{TimeMoE}$_{b}$} & \multicolumn{2}{l|}{\textbf{TimeMoE}$_{l}$} & \multicolumn{2}{l|}{\textbf{Timer-XL}} & \multicolumn{2}{l|}{\textbf{MOIRAI}$_{s}$} & \multicolumn{2}{l|}{\textbf{MOIRAI}$_{b}$} & \multicolumn{2}{l|}{\textbf{MOIRAI}$_{l}$} & \multicolumn{2}{l|}{\textbf{Chronos}$_{s}$} & \multicolumn{2}{l|}{\textbf{Chronos}$_{b}$} & \multicolumn{2}{l|}{\textbf{TimesFM}} & \multicolumn{2}{l}{\textbf{Moment}} \\
\multicolumn{2}{l}{} & \multicolumn{2}{|c|}{\textbf{(Ours)}}  & \multicolumn{2}{c|}{\citeyearpar{liu2025sundial}} & \multicolumn{2}{c|}{\citeyearpar{shi2024timemoe}} & \multicolumn{2}{c|}{\citeyearpar{shi2024timemoe}} & \multicolumn{2}{c|}{\citeyearpar{liu2025timerxl}} & \multicolumn{2}{c|}{\citeyearpar{woo2024moirai}} & \multicolumn{2}{c|}{\citeyearpar{woo2024moirai}} & \multicolumn{2}{c|}{\citeyearpar{woo2024moirai}} & \multicolumn{2}{c|}{\citeyearpar{ansari2024chronos}} & \multicolumn{2}{c|}{\citeyearpar{ansari2024chronos}} & \multicolumn{2}{c|}{\citeyearpar{timesfm}} & \multicolumn{2}{c}{\citeyearpar{goswami2024moment}} \\

\cmidrule(r){3-4} \cmidrule(r){5-6} \cmidrule(r){7-8} \cmidrule(r){9-10} \cmidrule(r){11-12} \cmidrule(r){13-14} \cmidrule(r){15-16} \cmidrule(r){17-18} \cmidrule(r){19-20} \cmidrule(r){21-22} \cmidrule(r){23-24} \cmidrule(r){25-26}
\multicolumn{2}{l|}{{Metrics}} & MSE & MAE & MSE & MAE & MSE & MAE & MSE & MAE & MSE & MAE & MSE & MAE & MSE & MAE & MSE & MAE & MSE & MAE & MSE & MAE & MSE & MAE & MSE & MAE \\
\midrule
\multirow{5}{*}{\rotatebox{90}{ETTh1}} & 96  & 0.365 & 0.397 & \textbf{0.348} & \textbf{0.385} & \textbf{0.357} & \textbf{0.381} & \textbf{0.350} & \textbf{0.382} & 0.369 & 0.391 & 0.401 & 0.402 & 0.376 & 0.392 & 0.381 & 0.388 & 0.466 & 0.409 & 0.440 & 0.393 & 0.414 & 0.404 & 0.688 & 0.557 \\
& 192 & \textbf{0.393} & \textbf{0.414} & \textbf{0.393} & 0.418 & \textbf{0.384} & \textbf{0.404} & \textbf{0.388} & \textbf{0.412} & 0.405 & \textbf{0.413} & 0.435 & 0.421 & 0.412 & \textbf{0.413} & 0.434 & 0.415 & 0.530 & 0.450 & 0.492 & 0.426 & 0.465 & 0.434 & 0.688 & 0.560 \\
& 336 & \textbf{0.410} & \textbf{0.423} & 0.422 & 0.440 & \textbf{0.411} & 0.434 & \textbf{0.411} & \textbf{0.430} & \textbf{0.418} & \textbf{0.423} & 0.438 & 0.434 & 0.433 & 0.428 & 0.485 & 0.445 & 0.570 & 0.486 & 0.500 & 0.462 & 0.503 & 0.456 & 0.675 & 0.563 \\
& 720 & \textbf{0.422} & \textbf{0.443} & 0.481 & 0.493 & 0.449 & 0.477 & \textbf{0.427} & \textbf{0.455} & \textbf{0.423} & \textbf{0.441} & 0.439 & 0.454 & 0.447 & \textbf{0.444} & 0.611 & 0.510 & 0.615 & 0.543 & 0.882 & 0.591 & 0.511 & 0.481 & 0.683 & 0.585 \\
 \rowcolor{tabhighlight}\cellcolor{white}& AVG & \textbf{0.395} & \textbf{0.419} & 0.411 & 0.434 & \textbf{0.400} & \textbf{0.424} & \textbf{0.394} & \textbf{0.419} & \textbf{0.404} & \textbf{0.417} & 0.428 & 0.427 & 0.417 & 0.419 & 0.480 & 0.439 & 0.545 & 0.472 & 0.591 & 0.468 & 0.473 & 0.443 & 0.683 & 0.566 \\
\midrule
\multirow{5}{*}{\rotatebox{90}{ETTh2}} & 96  & 0.284 & 0.342 & \textbf{0.271} & \textbf{0.333} & 0.305 & 0.359 & 0.302 & 0.354 & \textbf{0.283} & \textbf{0.342} & 0.297 & \textbf{0.336} & 0.294 & \textbf{0.330} & 0.296 & \textbf{0.330} & 0.307 & 0.356 & 0.308 & 0.343 & 0.315 & 0.349 & 0.342 & 0.396 \\
& 192 & \textbf{0.351} & 0.384 & 0.327 & \textbf{0.376} & \textbf{0.351} & 0.386 & 0.364 & 0.385 & \textbf{0.340} & \textbf{0.379} & 0.368 & 0.381 & 0.365 & \textbf{0.375} & 0.361 & \textbf{0.371} & 0.376 & 0.401 & 0.384 & 0.392 & 0.388 & 0.395 & 0.354 & 0.402 \\
& 336 & \textbf{0.368} & 0.405 & 0.354 & 0.402 & 0.391 & 0.418 & 0.417 & 0.425 & \textbf{0.366} & \textbf{0.400} & 0.370 & \textbf{0.393} & 0.376 & \textbf{0.390} & 0.390 & \textbf{0.390} & 0.408 & 0.431 & 0.429 & 0.430 & 0.422 & 0.427 & 0.356 & 0.407 \\
& 720 & \textbf{0.382} & \textbf{0.423} & \textbf{0.381} & 0.435 & 0.419 & 0.454 & 0.537 & 0.496 & \textbf{0.397} & \textbf{0.431} & 0.411 & 0.426 & 0.416 & 0.433 & 0.423 & 0.418 & 0.604 & 0.533 & 0.501 & 0.477 & 0.443 & 0.454 & \textbf{0.395} & \textbf{0.434} \\
\rowcolor{tabhighlight}\cellcolor{white}
& AVG & \textbf{0.346} & \textbf{0.389} & \textbf{0.333} & \textbf{0.387} & 0.366 & 0.404 & 0.405 & 0.415 & \textbf{0.347} & \textbf{0.388} & 0.361 & \textbf{0.384} & 0.362 & \textbf{0.382} & 0.367 & \textbf{0.377} & 0.424 & 0.430 & 0.405 & 0.410 & 0.392 & 0.406 & 0.361 & 0.409 \\
\midrule
\multirow{5}{*}{\rotatebox{90}{ETTm1}} 
& 96  & 0.369 & 0.380 & \textbf{0.280} & \textbf{0.334} & 0.338 & 0.368 & \textbf{0.309} & \textbf{0.357} & \textbf{0.317} & \textbf{0.356} & 0.418 & 0.392 & 0.363 & \textbf{0.356} & 0.380 & 0.361 & 0.511 & 0.423 & 0.454 & 0.408 & 0.361 & 0.370 & 0.654 & 0.527 \\
& 192 & 0.406 & 0.401 & \textbf{0.321} & \textbf{0.366} & \textbf{0.353} & 0.388 & \textbf{0.346} & \textbf{0.381} & 0.358 & \textbf{0.381} & 0.431 & 0.405 & 0.388 & \textbf{0.375} & 0.412 & 0.383 & 0.618 & 0.485 & 0.567 & 0.477 & 0.414 & 0.405 & 0.662 & 0.532 \\
& 336 & 0.436 & 0.419 & \textbf{0.350} & \textbf{0.389} & \textbf{0.381} & 0.413 & \textbf{0.373} & 0.408 & 0.386 & \textbf{0.401} & 0.433 & 0.412 & 0.416 & \textbf{0.392} & 0.416 & \textbf{0.392} & 0.683 & 0.524 & 0.662 & 0.525 & 0.445 & 0.429 & 0.672 & 0.537 \\
& 720 & 0.482 & 0.441 & \textbf{0.394} & \textbf{0.418} & 0.504 & 0.493 & 0.475 & 0.477 & \textbf{0.430} & \textbf{0.431} & 0.462 & 0.432 & \textbf{0.460} & \textbf{0.418} & 0.462 & \textbf{0.420} & 0.748 & 0.566 & 0.900 & 0.591 & 0.512 & 0.471 & 0.692 & 0.551 \\
\rowcolor{tabhighlight}\cellcolor{white}
& AVG & 0.423 & 0.410 & \textbf{0.336} & \textbf{0.377} & 0.394 & 0.415 & \textbf{0.376} & 0.405 & \textbf{0.373} & 0.392 & 0.436 & 0.410 & 0.406 & \textbf{0.385} & 0.422 & \textbf{0.391} & 0.640 & 0.499 & 0.645 & 0.500 & 0.433 & 0.418 & 0.670 & 0.536 \\
\midrule
\multirow{5}{*}{\rotatebox{90}{ETTm2}} 
& 96  & \textbf{0.186} & 0.272 & \textbf{0.170} & \textbf{0.256} & 0.201 & 0.291 & 0.197 & 0.286 & 0.198 & 0.288 & 0.214 & 0.288 & 0.205 & \textbf{0.273} & 0.211 & \textbf{0.274} & 0.209 & 0.291 & 0.199 & \textbf{0.274} & 0.197 & \textbf{0.271} & 0.260 & 0.335 \\
& 192 & \textbf{0.247} & 0.310 & \textbf{0.229} & \textbf{0.300} & 0.258 & 0.334 & 0.250 & 0.322 & \textbf{0.241} & \textbf{0.315} & 0.284 & 0.332 & 0.275 & 0.316 & 0.281 & 0.318 & 0.280 & 0.341 & 0.261 & 0.322 & 0.289 & 0.321 & 0.289 & 0.350 \\
& 336 & \textbf{0.305} & 0.346 & \textbf{0.281} & \textbf{0.337} & 0.324 & 0.373 & 0.337 & 0.375 & \textbf{0.286} & \textbf{0.348} & 0.331 & 0.362 & 0.329 & 0.350 & 0.341 & 0.355 & 0.354 & 0.390 & 0.326 & 0.366 & 0.360 & 0.366 & 0.324 & 0.369 \\
& 720 & \textbf{0.396} & 0.400 & \textbf{0.351} & \textbf{0.387} & 0.488 & 0.464 & 0.480 & 0.461 & \textbf{0.375} & \textbf{0.402} & 0.402 & 0.408 & 0.437 & 0.411 & 0.485 & 0.428 & 0.553 & 0.499 & 0.455 & 0.439 & 0.462 & 0.430 & 0.394 & \textbf{0.409} \\
\rowcolor{tabhighlight}\cellcolor{white}
& AVG & \textbf{0.284} & \textbf{0.298} & \textbf{0.258} & 0.320 & 0.317 & 0.365 & 0.316 & 0.361 & \textbf{0.273} & \textbf{0.336} & 0.307 & 0.347 & 0.311 & 0.337 & 0.329 & 0.343 & 0.349 & 0.380 & 0.310 & 0.350 & 0.328 & 0.346 & 0.316 & 0.365 \\
\midrule
\multirow{5}{*}{\rotatebox{90}{Weather}} 
& 96  & 0.178 & 0.236 & \textbf{0.157} & \textbf{0.205} & \textbf{0.160} & \textbf{0.214} & \textbf{0.159} & \textbf{0.213} & 0.171 & 0.225 & 0.198 & 0.222 & 0.220 & 0.217 & 0.199 & 0.211 & 0.211 & 0.243 & 0.203 & 0.238 & - & - & 0.243 & 0.255 \\
& 192 & \textbf{0.227} & \textbf{0.279} & \textbf{0.205} & \textbf{0.251} & \textbf{0.210} & 0.260 & 0.215 & 0.266 & \textbf{0.221} & \textbf{0.271} & 0.247 & 0.265 & 0.271 & 0.259 & 0.246 & \textbf{0.251} & 0.263 & 0.294 & 0.256 & 0.290 & - & - & 0.278 & 0.329 \\
& 336 & \textbf{0.279} & 0.316 & \textbf{0.253} & \textbf{0.289} & \textbf{0.274} & \textbf{0.309} & 0.291 & 0.322 & \textbf{0.274} & \textbf{0.311} & 0.283 & 0.303 & 0.286 & \textbf{0.297} & \textbf{0.274} & \textbf{0.291} & 0.321 & 0.339 & 0.314 & 0.336 & - & - & 0.306 & 0.346 \\
& 720 & \textbf{0.341} & \textbf{0.360} & \textbf{0.320} & \textbf{0.336} & 0.418 & 0.405 & 0.415 & 0.400 & \textbf{0.356} & \textbf{0.370} & 0.373 & \textbf{0.354} & 0.373 & \textbf{0.354} & \textbf{0.337} & \textbf{0.340} & 0.404 & 0.397 & 0.397 & 0.396 & - & - & \textbf{0.350} & 0.374 \\
\rowcolor{tabhighlight}\cellcolor{white}
& AVG & \textbf{0.256} & \textbf{0.298} & \textbf{0.234} & \textbf{0.270} & 0.265 & 0.297 & 0.270 & 0.300 & \textbf{0.256} & \textbf{0.294} & 0.275 & 0.286 & 0.287 & 0.281 & \textbf{0.264} & \textbf{0.273} & 0.300 & 0.318 & 0.292 & 0.315 & - & - & 0.294 & 0.326 \\
\midrule
\multirow{5}{*}{\rotatebox{90}{Global}} 
& 96  & 0.229 & 0.348 & - & - & \textbf{0.211} & \textbf{0.343} & \textbf{0.210} & \textbf{0.342} & - & - & 0.227 & 0.354 & \textbf{0.224} & \textbf{0.351} & \textbf{0.224} & \textbf{0.351} & 0.234 & 0.361 & 0.230 & 0.355 & 0.255 & 0.375 & 0.363 & 0.472 \\
& 192 & \textbf{0.268} & \textbf{0.389} & - & - & \textbf{0.257} & \textbf{0.386} & \textbf{0.254} & \textbf{0.385} & - & - & 0.269 & 0.396 & \textbf{0.266} & 0.394 & 0.267 & 0.395 & 0.276 & 0.400 & 0.273 & 0.395 & 0.313 & 0.423 & 0.387 & 0.489 \\
& 336 & 0.320 & 0.427 & - & - & \textbf{0.281} & \textbf{0.405} & \textbf{0.267} & \textbf{0.395} & - & - & \textbf{0.292} & 0.419 & \textbf{0.296} & 0.420 & \textbf{0.291} & 0.417 & 0.314 & 0.431 & 0.324 & 0.434 & 0.362 & 0.460 & 0.430 & 0.517 \\
& 720 & 0.403 & 0.481 & - & - & 0.354 & 0.465 & \textbf{0.289} & \textbf{0.420} & - & - & \textbf{0.351} & \textbf{0.437} & 0.403 & 0.498 & \textbf{0.387} & \textbf{0.488} & 0.418 & 0.504 & 0.505 & 0.542 & 0.486 & 0.545 & 0.582 & 0.617 \\
\rowcolor{tabhighlight}\cellcolor{white}
& AVG & 0.305 & 0.411 & - & - & \textbf{0.275} & \textbf{0.400} & \textbf{0.255} & \textbf{0.385} & - & - & \textbf{0.285} & \textbf{0.409} & 0.297 & 0.416 & 0.292 & 0.413 & 0.311 & 0.424 & 0.333 & 0.431 & 0.354 & 0.451 & 0.440 & 0.524 \\
\midrule
\rowc\rowcolor{blue!10}
\multicolumn{2}{c|}{\scalebox{1.1}{\textbf{Counts}}} & \multicolumn{2}{c|}{\scalebox{1.1}{\textbf{36}}} & \multicolumn{2}{c|}{\scalebox{1.1}{\textbf{38}}} & \multicolumn{2}{c|}{\scalebox{1.1}{23}} & \multicolumn{2}{c|}{\scalebox{1.1}{28}} & \multicolumn{2}{c|}{\scalebox{1.1}{\textbf{40}}} & \multicolumn{2}{c|}{\scalebox{1.1}{9}} & \multicolumn{2}{c|}{\scalebox{1.1}{19}} & \multicolumn{2}{c|}{\scalebox{1.1}{20}} & \multicolumn{2}{c|}{\scalebox{1.1}{0}} & \multicolumn{2}{c|}{\scalebox{1.1}{1}} & \multicolumn{2}{c|}{\scalebox{1.1}{1}} & \multicolumn{2}{c|}{\scalebox{1.1}{4}} \\
\bottomrule[1.5pt]
\end{tabular}}
\label{tab:main_results}
\end{table*}

%% file: Table/ablation_short.tex
\begin{table*}[t]
\centering
\setlength{\abovecaptionskip}{0.15cm}
\caption{Ablation results averaged over horizons (96, 192, 336, 720). Each cell shows \textit{Avg MSE / Avg MAE}. Best (lower) per dataset and overall average is \textbf{bold}. See details in Appendix Table~\ref{tab:ablation_all}}
\resizebox{\linewidth}{!}{
\begin{tabular}{l|cc|cc|cc|cc|cc|cc}
\toprule[1.5pt]
\multirow{2}{*}{\textbf{Method}} 
& \multicolumn{2}{c|}{\textbf{ETTh1}} 
& \multicolumn{2}{c|}{\textbf{ETTh2}} 
& \multicolumn{2}{c|}{\textbf{ETTm1}} 
& \multicolumn{2}{c|}{\textbf{ETTm2}} 
& \multicolumn{2}{c|}{\textbf{Weather}}
& \multicolumn{2}{c}{\textbf{Average}} \\
\cmidrule(r){2-13}
& MSE & MAE & MSE & MAE & MSE & MAE & MSE & MAE & MSE & MAE & MSE & MAE \\
\midrule
Baseline                               & \textbf{0.415} & \textbf{0.435} & \textbf{0.352} & \textbf{0.392} & 0.632 & 0.514 & 0.293 & 0.347 & 0.266 & 0.310 & 0.392 & 0.400 \\
Baseline + LEV                         & 0.452 & 0.457 & 0.358 & 0.399 & \textbf{0.611} & \textbf{0.507} & 0.303 & 0.352 & \textbf{0.263} & 0.311 & 0.397 & 0.405 \\
Baseline + LEV + SAGE                  & 0.420 & 0.439 & 0.356 & 0.396 & 0.612 & \textbf{0.507} & \textbf{0.291} & \textbf{0.345} & \textbf{0.263} & 0.308 & \textbf{0.388} & \textbf{0.399} \\
Baseline + LEV + SAGE + SCRN           & 0.426 & 0.441 & 0.361 & 0.399 & 0.634 & 0.514 & 0.297 & 0.351 & 0.267 & 0.314 & 0.397 & 0.404 \\
Baseline + LEV + SAGE + SCAD           & 0.448 & 0.459 & 0.369 & 0.403 & 0.612 & 0.511 & 0.309 & 0.352 & 0.266 & \textbf{0.307} & 0.401 & 0.406 \\
\bottomrule[1.5pt]
\end{tabular}
}
\vspace{-5pt}
\label{tab:ablation_avg}
\end{table*}

%% file: Appendix/dataset.tex
\paragraph{Benchmark Details.} We evaluate the performance of various models for long-term forecasting across eight well-established datasets, including the Weather~\citep{wu2021autoformer}, Global Temp~\citep{wu2023corrformer}, and ETT datasets (ETTh1, ETTh2, ETTm1, ETTm2)~\citep{haoyietal-informer-2021}. A detailed description of each dataset is provided in Table \ref{tab:dataset}. Forecastability is calculated by one minus the entropy of Fourier decomposition of time series \cite{goerg2013forecastable}. A larger value indicates better predictability.

%% file: Table/seg.tex
\begin{table}[h]
  \caption{Sensitivity analysis on segment length. Results are reported in MSE/MAE.}
  \label{tab:segment_length}
  \centering
  \scriptsize
  \begin{tabular}{lccc}
    \toprule
    Dataset & Segment=8 & Segment=16 & Segment=48 \\
    \midrule
    ETTh1 & 0.3874/0.4158 & 0.3893/0.4131 & 0.3899/0.4169 \\
          & 0.4101/0.4298 & 0.4135/0.4280 & 0.4099/0.4298 \\
          & 0.4189/0.4388 & 0.4205/0.4361 & 0.4178/0.4379 \\
          & 0.4370/0.4633 & 0.4378/0.4612 & 0.4398/0.4638 \\
    \midrule
    ETTh2 & 0.2795/0.3446 & 0.2889/0.3460 & 0.2855/0.3461 \\
          & 0.3412/0.3820 & 0.3513/0.3854 & 0.3496/0.3859 \\
          & 0.3757/0.4075 & 0.3783/0.4088 & 0.3803/0.4109 \\
          & 0.3887/0.4286 & 0.3909/0.4292 & 0.3934/0.4309 \\
    \midrule
    ETTm1 & 0.6043/0.4951 & 0.5998/0.4910 & 0.6123/0.5048 \\
          & 0.6238/0.5068 & 0.6217/0.5074 & 0.6293/0.5163 \\
          & 0.6455/0.5182 & 0.6407/0.5203 & 0.6457/0.5253 \\
          & 0.6720/0.5335 & 0.6642/0.5356 & 0.6680/0.5400 \\
    \midrule
    ETTm2 & 0.2045/0.2966 & 0.2067/0.2949 & 0.2047/0.2949 \\
          & 0.2587/0.3284 & 0.2590/0.3265 & 0.2591/0.3273 \\
          & 0.3154/0.3620 & 0.3094/0.3571 & 0.3126/0.3591 \\
          & 0.4061/0.4168 & 0.3974/0.4106 & 0.4022/0.4137 \\
    \midrule
    Weather & 0.1907/0.2573 & 0.1932/0.2573 & 0.1865/0.2494 \\
            & 0.2350/0.2927 & 0.2395/0.2923 & 0.2343/0.2871 \\
            & 0.2827/0.3261 & 0.2849/0.3231 & 0.2820/0.3205 \\
            & 0.3530/0.3730 & 0.3468/0.3655 & 0.3464/0.3636 \\
    \bottomrule
  \end{tabular}
\end{table}

%% file: Table/exo_noise.tex
\begin{table}[h]
  \caption{Ablation study on the exogenous noise vector size ($d_{\text{exo}}$). 
  Results are reported in MSE/MAE.}
  \label{tab:exo_noise}
  \centering
  \scriptsize
  \begin{tabular}{lccccc}
    \toprule
    Dataset & exo\_noise=1 & exo\_noise=2 & exo\_noise=4 & exo\_noise=6 & exo\_noise=8 \\
    \midrule
    ETTh1 & 0.4087/0.4258 & 0.4100/0.4261 & 0.4286/0.4343 & 0.4747/0.4654 & 0.5022/0.4806 \\
          & 0.4275/0.4380 & 0.4279/0.4368 & 0.4492/0.4489 & 0.4928/0.4790 & 0.5182/0.4916 \\
          & 0.4348/0.4477 & 0.4323/0.4442 & 0.4579/0.4604 & 0.5000/0.4897 & 0.5241/0.5010 \\
          & 0.4494/0.4738 & 0.4464/0.4705 & 0.4704/0.4843 & 0.5219/0.5210 & 0.5355/0.5267 \\
    \midrule
    ETTh2 & 0.2890/0.3507 & 0.2997/0.3567 & 0.2946/0.3542 & 0.3055/0.3618 & 0.3052/0.3661 \\
          & 0.3498/0.3875 & 0.3603/0.3925 & 0.3545/0.3903 & 0.3656/0.3983 & 0.3560/0.3965 \\
          & 0.3795/0.4124 & 0.3865/0.4140 & 0.3878/0.4170 & 0.3918/0.4207 & 0.3789/0.4157 \\
          & 0.3897/0.4300 & 0.3926/0.4306 & 0.3940/0.4329 & 0.3984/0.4369 & 0.3996/0.4380 \\
    \midrule
    ETTm1 & 0.5835/0.4966 & 0.5760/0.4889 & 0.5750/0.4877 & 0.5437/0.4843 & 0.5951/0.5076 \\
          & 0.6090/0.5091 & 0.6016/0.5009 & 0.5986/0.5001 & 0.5675/0.4969 & 0.6171/0.5173 \\
          & 0.6266/0.5194 & 0.6240/0.5123 & 0.6214/0.5131 & 0.5864/0.5064 & 0.6354/0.5263 \\
          & 0.6523/0.5373 & 0.6505/0.5296 & 0.6500/0.5290 & 0.6126/0.5227 & 0.6604/0.5390 \\
    \midrule
    ETTm2 & 0.2057/0.2946 & 0.2068/0.2946 & 0.2087/0.2960 & 0.2102/0.2955 & 0.2126/0.3009 \\
          & 0.2631/0.3283 & 0.2604/0.3264 & 0.2650/0.3286 & 0.2673/0.3279 & 0.2634/0.3309 \\
          & 0.3205/0.3621 & 0.3138/0.3591 & 0.3240/0.3639 & 0.3223/0.3606 & 0.3140/0.3614 \\
          & 0.4104/0.4163 & 0.4032/0.4145 & 0.4126/0.4183 & 0.4092/0.4144 & 0.3997/0.4137 \\
    \midrule
    Weather & 0.1915/0.2547 & 0.1971/0.2615 & 0.1961/0.2644 & 0.1971/0.2603 & 0.2013/0.2644 \\
            & 0.2385/0.2915 & 0.2417/0.2958 & 0.2364/0.2941 & 0.2390/0.2910 & 0.2435/0.2973 \\
            & 0.2859/0.3234 & 0.2854/0.3259 & 0.2801/0.3228 & 0.2824/0.3204 & 0.2851/0.3259 \\
            & 0.3439/0.3611 & 0.3489/0.3676 & 0.3414/0.3636 & 0.3428/0.3599 & 0.3480/0.3688 \\
    \bottomrule
  \end{tabular}
\end{table}

%% file: Table/ablation.tex
\begin{table*}[h]
\centering
\setlength{\abovecaptionskip}{0.15cm}
\caption{Ablation study across five datasets. Each cell shows \textit{MSE / MAE}. Best (lower) per dataset and horizon is \textbf{bold}.}

\begin{tabular}{c|c|cccc}
\toprule[1.5pt]
\textbf{Dataset} & \textbf{Method} &
\textbf{96} & \textbf{192} & \textbf{336} & \textbf{720} \\
\midrule
\multirow{5}{*}{ETTh1}
& Baseline                      & \textbf{0.389} / \textbf{0.413} & \textbf{0.414} / \textbf{0.428} & \textbf{0.420} / \textbf{0.436} & \textbf{0.438} / \textbf{0.461} \\
& Baseline + LEV               & 0.429 / 0.434 & 0.449 / 0.449 & 0.458 / 0.460 & 0.470 / 0.484 \\
& Baseline + LEV + SAGE        & 0.395 / 0.416 & 0.419 / 0.432 & 0.427 / 0.442 & 0.440 / 0.466 \\
& Baseline + LEV + SAGE + SCRN & 0.403 / 0.419 & 0.423 / 0.432 & 0.432 / 0.443 & 0.445 / 0.471 \\
& Baseline + LEV + SAGE + SCAD & 0.422 / 0.438 & 0.445 / 0.452 & 0.452 / 0.461 & 0.472 / 0.487 \\
\midrule
\multirow{5}{*}{ETTh2}
& Baseline                      & \textbf{0.289} / \textbf{0.346} & \textbf{0.351} / \textbf{0.385} & \textbf{0.378} / \textbf{0.409} & \textbf{0.391} / \textbf{0.429} \\
& Baseline + LEV               & 0.295 / 0.354 & 0.355 / 0.390 & 0.388 / 0.417 & 0.394 / 0.433 \\
& Baseline + LEV + SAGE        & 0.292 / 0.349 & 0.354 / 0.388 & 0.385 / 0.414 & 0.395 / 0.432 \\
& Baseline + LEV + SAGE + SCRN & 0.297 / 0.355 & 0.360 / 0.391 & 0.390 / 0.416 & 0.397 / 0.434 \\
& Baseline + LEV + SAGE + SCAD & 0.302 / 0.355 & 0.372 / 0.397 & 0.404 / 0.424 & 0.399 / 0.435 \\
\midrule
\multirow{5}{*}{ETTm1}
& Baseline                      & 0.600 / 0.491 & 0.622 / 0.507 & 0.641 / 0.520 & 0.664 / 0.536 \\
& Baseline + LEV               & \textbf{0.575} / 0.488 & \textbf{0.599} / \textbf{0.500} & 0.621 / 0.513 & 0.650 / \textbf{0.529} \\
& Baseline + LEV + SAGE        & 0.578 / \textbf{0.485} & 0.602 / 0.501 & \textbf{0.621} / \textbf{0.513} & 0.648 / 0.530 \\
& Baseline + LEV + SAGE + SCRN & 0.597 / 0.491 & 0.621 / 0.506 & 0.647 / 0.520 & 0.671 / 0.537 \\
& Baseline + LEV + SAGE + SCAD & 0.576 / 0.492 & 0.603 / 0.506 & 0.624 / 0.516 & \textbf{0.645} / 0.529 \\
\midrule
\multirow{5}{*}{ETTm2}
& Baseline                      & 0.207 / 0.295 & 0.259 / 0.327 & 0.309 / 0.357 & 0.397 / 0.411 \\
& Baseline + LEV               & 0.209 / 0.296 & 0.265 / 0.329 & 0.324 / 0.364 & 0.413 / 0.418 \\
& Baseline + LEV + SAGE        & \textbf{0.203} / \textbf{0.291} & \textbf{0.256} / \textbf{0.324} & \textbf{0.309} / \textbf{0.356} & \textbf{0.397} / \textbf{0.410} \\
& Baseline + LEV + SAGE + SCRN & 0.206 / 0.297 & 0.260 / 0.328 & 0.315 / 0.362 & 0.407 / 0.417 \\
& Baseline + LEV + SAGE + SCAD & 0.210 / 0.294 & 0.272 / 0.329 & 0.331 / 0.364 & 0.422 / 0.420 \\
\midrule
\multirow{5}{*}{Weather}
& Baseline                      & 0.193 / 0.257 & 0.239 / 0.292 & 0.285 / 0.323 & 0.347 / 0.366 \\
& Baseline + LEV               & 0.196 / 0.264 & 0.236 / 0.294 & 0.280 / 0.323 & \textbf{0.341} / \textbf{0.364} \\
& Baseline + LEV + SAGE        & \textbf{0.189} / 0.255 & \textbf{0.234} / 0.290 & \textbf{0.280} / 0.322 & 0.346 / 0.366 \\
& Baseline + LEV + SAGE + SCRN & 0.196 / 0.264 & 0.239 / 0.297 & 0.284 / 0.328 & 0.347 / 0.369 \\
& Baseline + LEV + SAGE + SCAD & 0.192 / \textbf{0.252} & 0.238 / \textbf{0.288} & 0.284 / \textbf{0.320} & 0.351 / 0.366 \\
\bottomrule[1.5pt]
\end{tabular}

\label{tab:ablation_all}
\end{table*}

%% file: Table/topk.tex
\begin{table*}[h]
\centering
\setlength{\abovecaptionskip}{0.15cm}
\caption{MoE Top-$k$ (total experts $E{=}8$). Numbers are averaged over horizons (96/192/336/720). Each cell shows \textit{Avg MSE / Avg MAE}. Best (lower) per dataset and overall is \textbf{bold}.}

\begin{tabular}{l|cc|cc|cc|cc|cc|cc}
\toprule[1.5pt]
\multirow{2}{*}{\textbf{Top-$k$}} 
& \multicolumn{2}{c|}{\textbf{ETTh1}} 
& \multicolumn{2}{c|}{\textbf{ETTh2}} 
& \multicolumn{2}{c|}{\textbf{ETTm1}} 
& \multicolumn{2}{c|}{\textbf{ETTm2}} 
& \multicolumn{2}{c|}{\textbf{Weather}}
& \multicolumn{2}{c}{\textbf{Overall}} \\
\cmidrule(r){2-13}
& MSE & MAE & MSE & MAE & MSE & MAE & MSE & MAE & MSE & MAE & MSE & MAE \\
\midrule
\textbf{Top-$k$=1}
& 0.421 & 0.440 & 0.354 & 0.396 & \textbf{0.624} & 0.520 & 0.296 & 0.349 & 0.265 & 0.309 & 0.392 & 0.403 \\
\textbf{Top-$k$=2}
& 0.422 & 0.442 & 0.358 & 0.398 & 0.636 & 0.524 & 0.297 & 0.350 & 0.267 & 0.311 & 0.396 & 0.405 \\
\textbf{Top-$k$=4}
& \textbf{0.415} & \textbf{0.433} & \textbf{0.354} & \textbf{0.394} & 0.625 & \textbf{0.516} & \textbf{0.289} & \textbf{0.344} & \textbf{0.265} & \textbf{0.307} & \textbf{0.390} & \textbf{0.399} \\
\textbf{Top-$k$=6}
& \textbf{0.415} & \textbf{0.433} & \textbf{0.354} & \textbf{0.394} & 0.625 & \textbf{0.516} & \textbf{0.289} & \textbf{0.344} & \textbf{0.265} & \textbf{0.307} & \textbf{0.390} & \textbf{0.399} \\
\textbf{Top-$k$=8}
& \textbf{0.415} & \textbf{0.433} & \textbf{0.354} & \textbf{0.394} & 0.625 & \textbf{0.516} & \textbf{0.289} & \textbf{0.344} & \textbf{0.265} & \textbf{0.307} & \textbf{0.390} & \textbf{0.399} \\
\bottomrule[1.5pt]
\end{tabular}

\label{tab:moe_topk_avg}
\end{table*}